\documentclass[12pt, english]{article}

\usepackage{Report}
\usepackage[useregional]{datetime2}
\usepackage{amsmath,amssymb,amsfonts}
\usepackage{algorithmic}
\usepackage{graphicx}
\usepackage{pifont}
\usepackage{textcomp}
\usepackage{hyperref}
\usepackage{multirow}
\usepackage{balance}
\usepackage{indentfirst}
\usepackage{url}
\usepackage[export]{adjustbox}
\usepackage{comment}
\usepackage{color}
\usepackage{latexsym}
\usepackage{xurl}
\usepackage{mathtools}
\usepackage{caption}
\usepackage{subcaption}
\usepackage{mathrsfs}

\graphicspath{{Figures/}}
\bibliography{References}
\begin{document}
\renewcommand{\familydefault}{\rmdefault}

\begin{titlepage}
    \begin{center}
    {\fontsize{22}{32}\selectfont \bfseries Autonomy Oriented Digital Twins for Real2Sim2Real Autoware Deployment}
    \\\vspace{85pt}
    {\LARGE Technical Report} \\
    \vspace{20pt}
    \textbf{Authors} \\
    {Chinmay Vilas Samak \\
     Tanmay Vilas Samak}
    \vspace{50pt} \\
    Fall 2023
    \end{center}

    \vspace{80pt}

    \begin{figure}[h]
    \centering
    \includegraphics[width=0.5\linewidth]{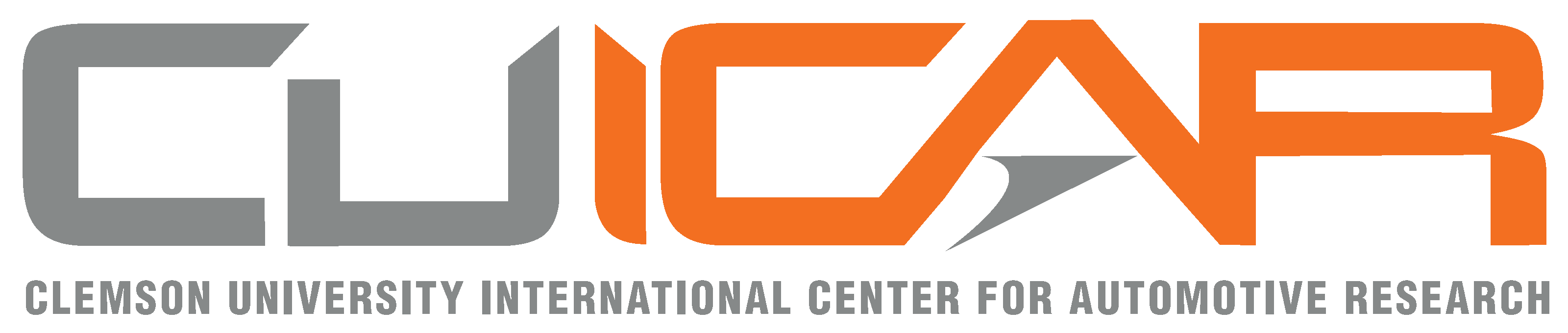}
    \end{figure}

    \vspace{10pt}

    \begin{table*}[h]
    \centering
    \begin{tabular}{l}
    \hline
    \begin{tabular}[c]{p{0.9\textwidth}}{\small This work is an outcome of the course AuE-8360 ``Scaled Autonomous Vehicles'' (Fall 2023) at Clemson University International Center for Automotive Research (CU-ICAR) handled by Dr. Venkat Krovi, Michelin Endowed SmartState Chair Professor of Vehicle Automation at CU-ICAR.}\\ \\ {\small This report was submitted to the Department of Automotive Engineering, CU-ICAR as a part of the course AuE-8360 ``Scaled Autonomous Vehicles''.}\end{tabular} \\ \hline
    \end{tabular}
    \end{table*}

    \pagebreak

    \section*{Disclaimer}\label{Disclaimer}
    \thispagestyle{empty}

    I certify that all the work and writing that I contributed to here is my own and not acquired from external sources. I have cited sources appropriately and paraphrased correctly. I have not shared my writing with other students (for individual assignments) and other students outside my group (for group project), nor have I acquired any written portion of this document from past or present students.

    \begin{flushright}
    \textbf{Authors}
    \end{flushright}

    \pagebreak

    \section*{Abstract}\label{Abstract}
    \thispagestyle{empty}

    Modeling and simulation of autonomous vehicles plays a crucial role in achieving enterprise-scale realization that aligns with technical, business and regulatory requirements. Contemporary trends in digital lifecycle treatment have proven beneficial to support simulation-based-design (SBD) as well as verification and validation (V\&V) of these complex systems. Although, the development of appropriate fidelity simulation models capable of capturing the intricate real-world physics and graphics (real2sim), while enabling real-time interactivity for decision-making, has remained a challenge. Nevertheless, recent advances in AI-based tools and workflows, such as online deep-learning algorithms leveraging live-streaming data sources, offer the tantalizing potential for real-time system-identification and adaptive modeling to simulate vehicle(s), environment(s), as well as their interactions. This transition from static/fixed-parameter ``virtual prototypes'' to dynamic/adaptable ``digital twins'' not only improves simulation fidelity and real-time factor, but can also support the development of online adaption/augmentation techniques that can help bridge the gap between simulation and reality (sim2real). In such a milieu, this work focuses on developing autonomy-oriented digital twins of vehicles across different scales and configurations to help support the streamlined development and deployment of Autoware Core/Universe stack, using a unified real2sim2real toolchain. Particularly, the core deliverable for this project was to integrate the Autoware stack with AutoDRIVE Ecosystem to demonstrate end-to-end task of mapping an unknown environment, recording a trajectory within the mapped environment, and autonomously tracking the pre-recorded trajectory to achieve the desired objective. This work discusses the development of vehicle and environment digital twins using AutoDRIVE Ecosystem, along with various application programming interfaces (APIs) and human-machine interfaces (HMIs) to connect with the same, followed by a detailed section on AutoDRIVE-Autoware integration. Furthermore, this study describes the first-ever off-road deployment of the Autoware stack, expanding the operational design domain (ODD) beyond on-road autonomous navigation.\\

    \noindent{\textbf{Keywords:} Autonomous Vehicles, Autoware, Digital Twins, Real2Sim, Sim2Real}

    \pagebreak

    \section*{Acknowledgment}\label{Acknowledgment}
    \thispagestyle{empty}

    At the commencement of this report, we would like to add a few words of appreciation for all those who have been a part of this project; directly or indirectly.
    
    We would like to express our immense gratitude towards Department of Automotive Engineering, Clemson University International Center for Automotive Research (CU-ICAR) for offering this course and for providing us with an excellent atmosphere for working on this project.
    
    We would also like to express deep and sincere gratitude to our faculty Dr. Venkat Krovi for his valuable guidance, consistent encouragement and timely help. We are also thankful of our peers in the course, who engaged in thoughtful conversations and discussions, which led to the cross-pollination of ideas.
    
    Finally, we are grateful to all the sources of information without which this project would be incomplete. It is due to their efforts and research that our report is more accurate and convincing.

    \begin{flushright}
    \textbf{Authors}
    \end{flushright}

    \pagebreak

    \thispagestyle{empty}
    \tableofcontents
    \thispagestyle{empty}
    
    \pagebreak
    
\end{titlepage}

\pagestyle{fancy}
\fancyhf{}
\setlength{\headheight}{30pt}
\renewcommand{\headrulewidth}{0.4pt}
\renewcommand{\footrulewidth}{0.4pt}
\lhead{AutoDRIVE-Autoware Integration}
\rhead{C.V. Samak and T.V. Samak}
\rfoot{\textbf{Page \thepage}}
\lfoot{}
\pagebreak


\fontsize{12}{20}\selectfont{

\hypertarget{Introduction}{%
\section{Introduction}\label{Introduction}}

\begin{figure}[htpb]
    \centering
    \includegraphics[width=\linewidth]{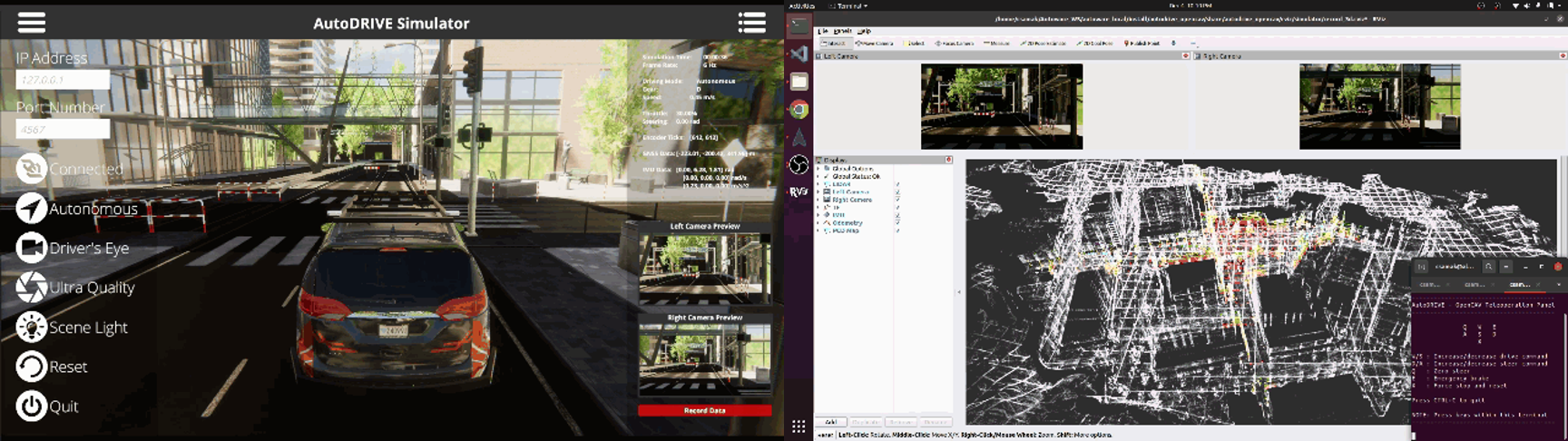}
    \caption{Project overview demonstrating the integration of AutoDRIVE Ecosystem with Autoware stack for the application of autonomous valet parking using the OpenCAV.}
    \label{fig: figure1}
\end{figure}

\hypertarget{Motivation}{%
\subsection{Motivation}\label{Motivation}}

Modeling and simulation of autonomous vehicles plays a crucial role in achieving enterprise-scale realization that aligns with technical, business and regulatory requirements. Contemporary trends in digital lifecycle treatment have proven beneficial to support simulation-based-design (SBD) as well as verification and validation of increasingly complex systems and system-of-systems. However, the development of appropriate fidelity simulation models capable of capturing the intricate real-world physics and graphics (real2sim), while enabling real-time interactivity for decision-making, has remained a challenge.

Autonomy-oriented simulations, as opposed to conventional simulations, must equally prioritize back-end physics and front-end graphics, which is crucial for realistic simulation of vehicle dynamics, sensor characteristics, and environmental physics. Additionally, the interconnect between vehicles, sensors, actuators and the environment, along with peer vehicles and infrastructure in a scene must be appropriately modeled. Most importantly, however, these simulations should allow real-time interfacing with software development framework(s) to support reliable verification and validation of autonomy algorithms.

Recent advances in AI-based tools and workflows, such as online deep-learning algorithms leveraging live-streaming data sources, offer the tantalizing potential for real-time system-identification and adaptive modeling to simulate vehicle(s), environment(s), as well as their interactions. This transition from static/fixed-parameter ``virtual prototypes'' to dynamic/adaptable ``digital twins'' not only improves simulation fidelity and real-time factor, but can also support the development of online adaption/augmentation techniques that can help bridge the gap between simulation and reality (sim2real).

However, seamlessly moving from reality to simulation and back to reality (real2sim2real) \cite{Real2Sim2Real} requires a sleek workflow in place. In such a milieu, this work focuses on developing autonomy-oriented digital twins of vehicles across different scales and configurations to help support the streamlined development and deployment of Autoware Core/Universe stack \cite{AutowareCore, AutowareUniverse, AutowareStack}, using a unified real2sim2real toolchain. Needless to say, these ``autonomy-oriented'' digital twins, as opposed to conventional simulation models, prioritize equal detailing of physics and graphics.

\hypertarget{Objectives}{%
\subsection{Objectives}\label{Objectives}}

The core deliverable of this project was integrating Autoware\footnote{\url{https://autoware.org}} \cite{Autoware} stack with AutoDRIVE Ecosystem\footnote{\url{https://autodrive-ecosystem.github.io}} \cite{AutoDRIVEEcosystem, AutoDRIVESimulator, AutoDRIVEReport, AutoDRIVESimulatorReport} to demonstrate the end-to-end task of mapping an unknown environment, recording a trajectory within the mapped environment, and autonomously tracking the pre-recorded trajectory to achieve the desired objective (Fig. \ref{fig: figure1}). Particularly, we demonstrate sim2real Autoware deployments using Nigel \cite{Nigel} and F1TENTH \cite{F1TENTH}, two small-scale autonomous vehicle platforms with unique qualities and capabilities. Additionally we demonstrate simulated Autoware deployments using mid-scale Hunter SE \cite{HunterSE} and full-scale OpenCAV \cite{OpenCAV} within simplistic and realistic scenarios. It is worth mentioning that this study describes the first-ever off-road deployment of the Autoware stack using the mid-scale vehicle platform, thereby expanding the operational design domain (ODD) of Autoware beyond on-road autonomous navigation.

As a precursor to Autoware deployments, this work discusses the development of vehicle and environment digital twins using AutoDRIVE Ecosystem, which span across different scales and operational design domains. The development of these autonomy-oriented digital twins was, therefore, the secondary objective of this project. This step involved developing geometric as well as dynamics models of vehicles and calibrating them against their real-world counterparts. Additionally, developing physics-based models for interoceptive as well as exteroceptive sensors and actuators was accomplished based on their respective datasheets. Finally, creating physically and graphically realistic on-road and off-road environments across scales marked the completion of this objective.

The tertiary objective of this project was to develop cross-platform application programming interfaces (APIs) and human-machine interfaces (HMIs) to connect with AutoDRIVE Ecosystem, which would aid in AutoDRIVE-Autoware integration. This objective, in conjunction with the secondary objective enabled the realization of the primary objective of developing a streamlined real2sim2real Autoware development framework with deployment demonstrations across varying scales and ODDs.

\hypertarget{Applications}{%
\subsection{Applications}\label{Applications}}

Following is a brief summary of the potential applications and scope of this project, which align well with the different ODDs and use cases of Autoware Foundation \cite{AutowareOverview}:

\begin{itemize}
    \item \textbf{Autonomous Valet Parking (AVP)}: Mapping of a parking lot, localization within the created map and autonomous driving within the parking lot.
    \item \textbf{Cargo Delivery}: Autonomous mobile robots for the transport of goods between multiple points or last-mile delivery.
    \item \textbf{Racing}: Autonomous racing using small-scale (e.g. F1TENTH) and full-scale (e.g. Indy Autonomous Challenge \cite{IAC}) vehicles running the Autoware stack.
    \item \textbf{Robo-Bus/Shuttle}: Fully autonomous (Level 4) buses and shuttles operating on public roads with predefined routes and stops.
    \item \textbf{Robo-Taxi}: Fully autonomous (Level 4) taxis operating in dense urban environments to pick-up and drop passengers from point-A to point-B.
    \item \textbf{Off-Road Exploration}: Although the Autoware Foundation has not yet proposed such an ODD, this study describes the first-ever off-road deployment of the Autoware stack using the mid-scale vehicle platform, thereby expanding the operational design domain (ODD) of Autoware beyond on-road autonomous navigation. Such off-road deployments could be applied for agricultural, military or extra-terrestrial applications.
\end{itemize}

\pagebreak
\hypertarget{Project Management}{%
\section{Project Management}\label{Project Management}}

\hypertarget{Materials and Methods}{%
\subsection{Materials and Methods}\label{Materials and Methods}}

Following is a brief summary of the tools (Fig. \ref{fig: figure1}) that went into designing and implementing this project:

\begin{itemize}
    \item \textbf{Digital Twin Ecosystem:} \href{https://autodrive-ecosystem.github.io}{AutoDRIVE Ecosystem}
    \item \textbf{Vehicle Platforms:} \href{https://youtu.be/YFQzyfXV6Rw?feature=shared}{Nigel}, \href{https://youtu.be/Rq7Wwcwn1uk?si=_ODExkHBopsQszrU}{F1TENTH}, \href{https://global.agilex.ai/chassis/9}{Hunter SE} and \href{https://youtu.be/YIZz_8rLgZQ?si=6Z6LWxWSTyS3Uk8u}{OpenCAV}
    \item \textbf{Software Stack:} \href{https://github.com/autowarefoundation/autoware.universe/tree/galactic}{Autoware Universe - ROS 2 Galactic}
    \item \textbf{Programming:} \href{https://www.python.org}{Python} and \href{https://isocpp.org}{C++}
\end{itemize}

\begin{figure}[htpb]
    \centering
    \includegraphics[width=\linewidth]{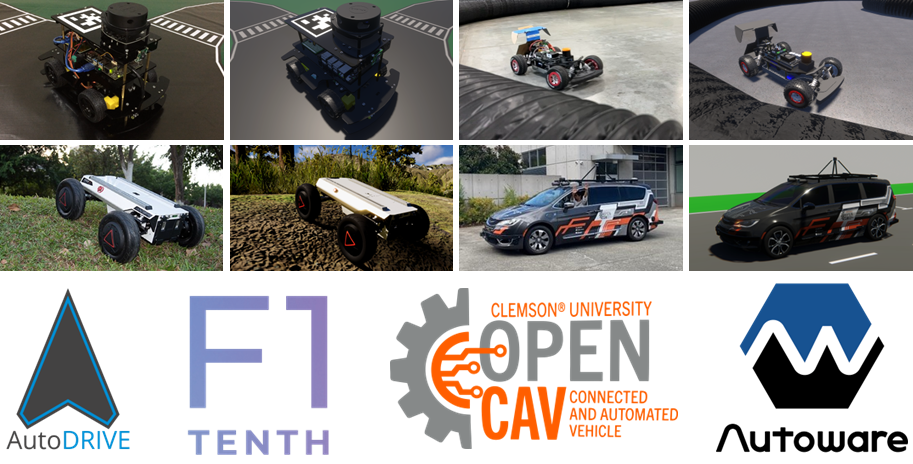}
    \caption{Project tools: AutoDRIVE Ecosystem employed to develop autonomy-oriented digital twins of small-scale (Nigel and F1TENTH), mid-scale (Hunter SE) and full-scale (OpenCAV) vehicles for deploying the Autoware stack.}
    \label{fig: figure2}
\end{figure}

The vehicles used for this project varied across scales (Fig. \ref{fig: figure2}). Small-scale platforms included Nigel (1:14 scale) and F1TENTH (1:10 scale), mid-scale platforms included Hunter SE (1:5 scale) and full-scale platforms included the OpenCAV (1:1 scale). Consequently, the sensor suite for Nigel and F1TENTH\footnote{The virtual F1TENTH was modeled with all sensors modules but only a subset of them were used for physical deployments} included small-scale sensors such as throttle and steering sensors, incremental encoders, indoor-positioning system (IPS), inertial measurement unit (IMU), RGB cameras and a 2D LIDAR. Alternatively, the sensor suite for Hunter SE and OpenCAV included different variants of 3D LIDARs in addition to similar virtual sensors. The actuators for small and mid-scale vehicles comprised throttle and steering actuators, which provided driving and steering torques to the respective wheels. Alternatively, the OpenCAV had a detailed powertrain model and exhaustive control inputs (throttle, steering, brake, handbrake) corresponding to its real-world counterpart. Finally, each vehicle was simulated in environment(s) appropriate for its scale and operational design domain.

For digital twinning, we adopted \href{https://github.com/Tinker-Twins/AutoDRIVE/tree/AutoDRIVE-Simulator}{AutoDRIVE Simulator}, a high-fidelity simulation system for autonomy-oriented applications. The simulation models were calibrated and benchmarked against their real-world counterparts for perception and dynamics reliability. The same simulation framework was also used to develop various HMIs that could directly interface with the virtual vehicles.

Core API development along with Autoware integration for all the virtual/real vehicles was accomplished using \href{https://github.com/Tinker-Twins/AutoDRIVE/tree/AutoDRIVE-Devkit}{AutoDRIVE Devkit}. The developed APIs could interface the virtual/real vehicles with Python, C++, ROS \cite{ROS1}, ROS 2 \cite{ROS2} or the Autoware stack. Additionally, the said framework also facilitated the development of API-mediated HMIs for the virtual as well as physical vehicles. 

Finally, we employed \href{https://github.com/Tinker-Twins/AutoDRIVE/tree/AutoDRIVE-Testbed}{AutoDRIVE Testbed} for sim2real deployment of Autoware demonstrations, which was integrated with \href{https://github.com/Tinker-Twins/AutoDRIVE/tree/AutoDRIVE-Devkit}{AutoDRIVE Devkit} deployed on Nigel and F1TENTH.

\hypertarget{Work Breakdown Structure}{%
\subsection{Work Breakdown Structure}\label{Work Breakdown Structure}}

For our convenience, we divided the project into 3 phases, each focusing on a different scale of autonomous vehicle platform for developing vehicle and environment digital twins, integrating various APIs and HMIs, and demonstrating an end-to-end autonomy deployment using the Autoware stack. The first phase of the project was set as the core deliverable and the rest of them as ``ambitious'' extensions subject to time-availability:
\begin{itemize}
    \item \textbf{Phase 1: Small-Scale Deployments:}
    \begin{itemize}
    \item Develop digital twins of Nigel and F1TENTH vehicles
    \item Develop digital twins of small-scale environments suitable for respective vehicles
    \item Develop APIs and HMIs to connect with virtual and real vehicles
    \item Integrate Autoware stack with virtual and real vehicles
    \item Demonstrate sim2real deployment of Autoware stack on both the vehicles
    \end{itemize}

    \item \textbf{Phase 2: Mid-Scale Deployments:}
    \begin{itemize}
    \item Develop digital twins of Hunter SE and Husky robots
    \item Develop digital twins of mid-scale environments suitable for respective vehicles
    \item Develop APIs and HMIs to connect with virtual and real vehicles
    \item Integrate Autoware stack with virtual and real Hunter SE
    \item Demonstrate simulated deployment of Autoware stack on Hunter SE
    \end{itemize}

    \item \textbf{Phase 3: Full-Scale Deployments:}
    \begin{itemize}
    \item Develop digital twins of OpenCAV and RZR vehicles
    \item Develop digital twins of full-scale environments suitable for respective vehicles
    \item Develop APIs and HMIs to connect with virtual and real vehicles
    \item Integrate Autoware stack with virtual and real OpenCAV
    \item Demonstrate simulated deployment of Autoware stack on OpenCAV
    \end{itemize}
\end{itemize}

\hypertarget{Project Timeline}{%
\subsection{Project Timeline}\label{Project Timeline}}

As mentioned earlier, we split this project into three phases viz. small-scale, mid-scale and full-scale Autoware deployments. Additionally, logistical tasks such as continuous documentation, preparation of slide decks, presentation rehearsals, report writing, etc. had to be accounted for in the overall project plan and strictly followed to ensure successful completion of the project.

\begin{figure}[htpb]
    \centering
    \includegraphics[width=\linewidth]{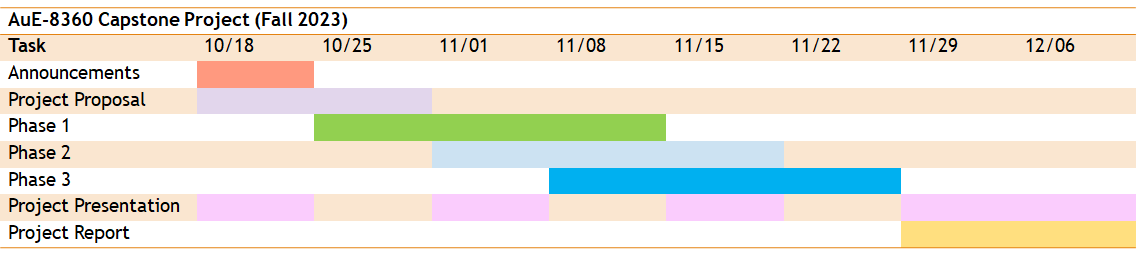}
    \caption{Project timeline Gantt chart.}
    \label{fig: figure3}
\end{figure}

Given the overall timeline from 10/18/2023 to 12/13/2023 (end of Fall 2023 semester), the milestones and deliverables of the project were planned as indicated in (Fig. \ref{fig: figure3}), which depicts the Gantt chart. It is to be noted that factors such as known parallel commitments were weighed in while preparing the timeline to enable practical adherence.

\hypertarget{Responsibility Assignment}{%
\subsection{Responsibility Assignment}\label{Responsibility Assignment}}

We divided the project into a set of tasks and assigned a subset of the team members with a primary and secondary responsibility of accomplishing each task as per the planned deadlines. Upon successful implementation of the project, albeit barebone, we came together for overall project optimization, organization and documentation.

\begin{itemize}
    \item \textbf{Primary responsibilities for Tanmay:}
    \begin{itemize}
    \item Develop digital twins of Hunter SE and Husky robots
    \item Develop digital twins of OpenCAV and RZR vehicles
    \item Develop digital twins of mid-scale environments suitable for respective vehicles
    \item Develop digital twins of full-scale environments suitable for respective vehicles
    \item Design and develop physical prototype of the Nigel (4WD4WS) vehicle
    \item Demonstrate sim/real deployment of Autoware stack on all the vehicles
    \end{itemize}
    
    \item \textbf{Primary responsibilities for Chinmay:}
    \begin{itemize}
    \item Develop digital twins of Nigel and F1TENTH vehicles
    \item Develop digital twins of small-scale environments suitable for respective vehicles
    \item Setup and calibrate physical prototype of the F1TENTH vehicle
    \item Develop APIs and HMIs to connect with virtual and real vehicles
    \item Integrate Autoware stack with virtual and real vehicles
    \item Project presentations, report and documentation
    \end{itemize}
\end{itemize}

It is to be noted that ``responsibility'' does not directly indicate ``contribution''. The team members have contributed equally and have no conflict of interest to declare.

\pagebreak
\hypertarget{Digital Twin Implementation}{%
\section{Digital Twin Implementation}\label{Digital Twin Implementation}}

Traditionally, automotive industry has long practiced the gradual transition from virtual, to hybrid, to physical phases within an X-in-the-loop (XIL; X = model, software, processor, hardware, vehicle) framework. Furthermore, recent modeling \& simulation methodologies such as simulation-as-a-service (SAAS) complemented with simulation-to-reality (sim2real) transfer show promise in facilitating parallelized scenario-based validation to enable robust system development and comprehensive corner-case analysis. However, the lack of physically realistic simulation of perception characteristics, system dynamics and agent-environment interactions, along with associated uncertainties, restricts the scalability and reliability of simulation-based verification. 

In such a milieu, digital twins have emerged as potentially viable tools to improve simulation fidelity and to develop adaption/augmentation techniques that can help bridge the sim2real gap. The following sections delve into the development of physically and graphically accurate vehicle and environment digital twins, and also discuss the integration of these with APIs and HMIs for developing autonomy-oriented applications.

\hypertarget{Vehicle Digital Twins}{%
\subsection{Vehicle Digital Twins}\label{Vehicle Digital Twins}}

\begin{figure}[t]
    \centering
    \includegraphics[width=\linewidth]{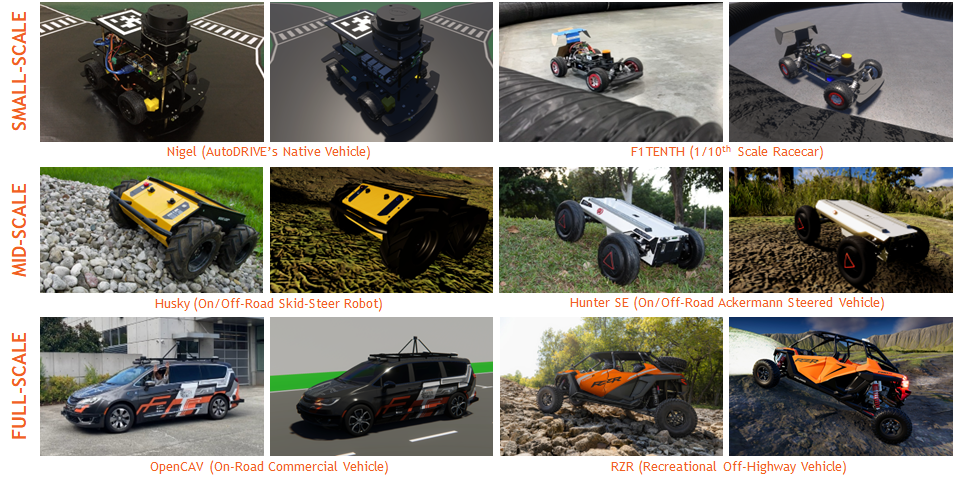}
    \caption{Autonomy-oriented vehicle digital twins across scales: Nigel and F1TENTH (small-scale), Husky and Hunter SE (mid-scale), and OpenCAV and RZR (full-scale) platforms for on/off-road autonomy.}
    \label{fig: figure4}
\end{figure}

As described earlier, we leveraged AutoDRIVE Simulator \cite{AutoDRIVESimulator, AutoDRIVESimulatorReport} to develop digital twin models of six different vehicles, across different scales and ODDs (Fig. \ref{fig: figure4}). These included small-scale (Nigel and F1TENTH), mid-scale (Husky and Hunter SE) and full-scale (OpenCAV and RZR) vehicles targeted towards on-road as well as off-road autonomy. It is to be noted that Husky and RZR fall purely under off-road ODD, and since this project (as well as Autoware) primarily targets on-road autonomy, these platforms have not been discussed elaborately in this report. Consequently, the following sections describe the digital twins of Nigel, F1TENTH, Hunter SE and OpenCAV in detail. This elucidation covers details at component, sub-system, system, all the way up to complex system-of-systems level models and their interconnects.

From a computational perspective, the said simulation framework was developed modularly using object-oriented programming (OOP) constructs. Additionally, the simulator took advantage of CPU multi-threading as well as GPU instancing (if available) to efficiently handle the workload, while providing cross-platform support.

\subsubsection{Vehicle Models}
\label{Vehicle Models}

\begin{figure}[t]
    \centering
    \includegraphics[width=\linewidth]{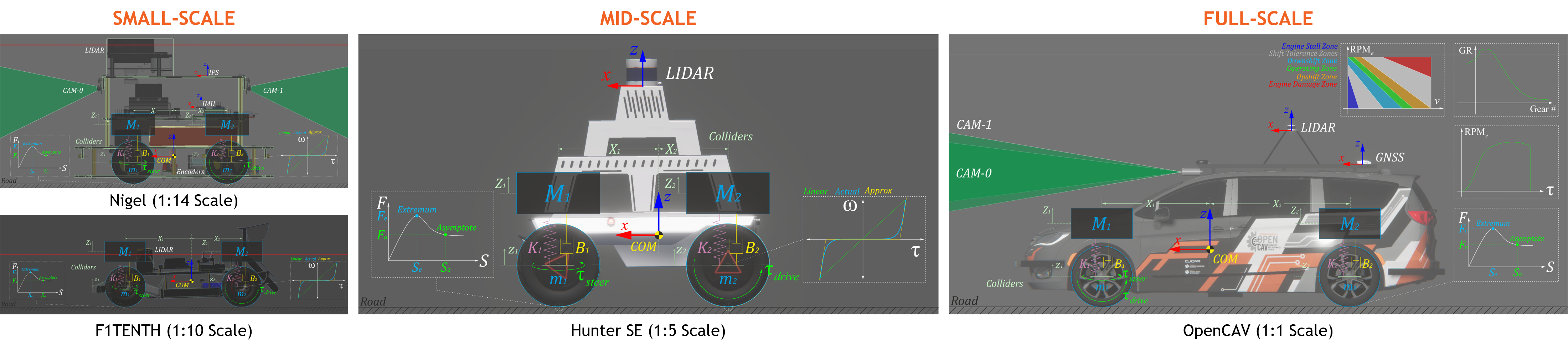}
    \caption{Simulation of vehicle dynamics, sensors and actuators for Nigel and F1TENTH digital twins.}
    \label{fig: figure5}
\end{figure}

The vehicles (refer Fig. \ref{fig: figure5}) are conjunctly modeled using sprung-mass ${^iM}$ and rigid-body representations. Here, the total mass $M=\sum{^iM}$, center of mass, $X_{COM} = \frac{\sum{{^iM}*{^iX}}}{\sum{^iM}}$ and moment of inertia $I_{COM} = \sum{{^iM}*{^iX^2}}$, serve as the linkage between these two representations, where ${^iX}$ represents the coordinates of the sprung masses. Each vehicle's wheels are also modeled as rigid bodies with mass $m$, experiencing gravitational and suspension forces: ${^im} * {^i{\ddot{z}}} + {^iB} * ({^i{\dot{z}}}-{^i{\dot{Z}}}) + {^iK} * ({^i{z}}-{^i{Z}})$.

\paragraph{Powertrain Dynamics}
\label{Powertrain Dynamics}

For small and mid-scale vehicles, which usually implement an electric motor for propulsion, the front/rear/all wheels are driven by applying a torque ${^i\tau_{drive}} = {^iI_w}*{^i\dot{\omega}_w}$, where ${^iI_w} = \frac{1}{2}*{^im_w}*{^i{r_w}^2}$ represents the moment of inertia, $^i\dot{\omega}_w$ is the angular acceleration, $^im_w$ is the mass, and $^ir_w$ is the radius of the $i$-th wheel. The actuation delays can also be modeled by splitting the torque profile into multiple segments based on operating conditions. For full-scale vehicles, however, the powertrain comprises an engine, transmission and differential. The engine is modeled based on its torque-speed characteristics. The engine RPM is updated smoothly based on its current value $RPM_e$, the idle speed $RPM_i$, average wheel speed $RPM_w$, final drive ratio $FDR$, current gear ratio $GR$, and the vehicle velocity $v$. The update can be expressed as $RPM_e := \left[RPM_i + \left(|RPM_w| \times FDR * GR\right)\right]_{(RPM_e,v)}$ where, $[\mathscr{F}]_x$ denotes evaluation of $\mathscr{F}$ at $x$. The total torque generated by the powertrain is computed as $\tau_{\text{total}} = \left[\tau_e\right]_{RPM_e} * \left[GR\right]_{G_\#} * FDR * T * \mathscr{A}$. Here, $\tau_e$ is the engine torque, $T$ is the throttle input, and $\mathscr{A}$ is a non-linear smoothing operator which increases the vehicle acceleration based on the throttle input. The automatic transmission decides to upshift/downshift the gears based on the transmission map of a given vehicle. This keeps the engine RPM in a good operating range for a given speed: $RPM_e = \frac{{v_{\text{MPH}} * 5280 * 12}}{{60 * 2 * \pi * R_{\text{tire}}}} * FDR * GR$. It is to be noted that while shifting the gears, the total torque produced by the powertrain is set to zero to simulate the clutch disengagement. It is also noteworthy that the auto-transmission is put in neutral gear once the vehicle is in standstill condition and parking gear if handbrakes are engaged in standstill condition. Additionally, switching between drive and reverse gears requires that the vehicle first be in the neutral gear to allow this transition. The total torque $\tau_\text{total}$ from the drivetrain is divided to the wheels based on the drive configuration of the vehicle:
$
\tau_{\text{out}} = \begin{cases}
\frac{\tau_{\text{total}}}{2} & \text{if FWD/RWD} \\
\frac{\tau_{\text{total}}}{4} & \text{if AWD}
\end{cases}
$
The torque transmitted to wheels $\tau_w$ is modeled by dividing the output torque $\tau_\text{out}$ to the left and right wheels based on the steering input. The left wheel receives a torque amounting to $^{L}\tau_{w} = \tau_{\text{out}} * (1 - \tau_{\text{drop}} * |\delta^{-}|)$, while the right wheel receives a torque equivalent to $^{R}\tau_{w} = \tau_{\text{out}} * (1 - \tau_{\text{drop}} * |\delta^{+}|)$. Here, $\tau_\text{drop}$ is the torque-drop at differential and $\delta^{\pm}$ indicates positive and negative steering angles, respectively. The value of $(\tau_{\text{drop}} * |\delta^{\pm}|)$ is clamped between $[0,0.9]$.

\paragraph{Brake Dynamics}
\label{Brake Dynamics}

The driving actuators for small and mid-scale vehicles simulate braking torque by applying a holding torque in idle conditions, i.e., ${^i\tau_\text{brake}} = {^i\tau_\text{idle}}$. For full-scale vehicles, the braking torque is modeled as ${^i\tau_\text{brake}} = \frac{{^iM}*v^2}{2*D_\text{brake}}*R_b$ where $R_b$ is the brake disk radius and $D_\text{brake}$ is the braking distance at 60 MPH, which can be obtained from physical vehicle tests. This braking torque is applied to the wheels based on the type of brake input: for combi-brakes, this torque is applied to all the wheels, and for handbrakes, it is applied to the rear wheels only.

\paragraph{Steering Dynamics}
\label{Steering Dynamics}

The steering mechanism operates by employing a steering actuator, which applies a torque $\tau_{\text{steer}}$ to achieve the desired steering angle $\delta$ with a smooth rate $\dot{\delta}$, without exceeding the steering limits $\pm \delta_\text{lim}$. The rate at which the vehicle steers is governed by its speed $v$ and steering sensitivity $\kappa_\delta$, and is represented as $\dot{\delta} = \kappa_\delta + \kappa_v * \frac{v}{v_\text{max}}$. Here, $\kappa_v$ is the speed-dependency factor of the steering mechanism. Finally, the individual angle for left $\delta_l$ and right $\delta_r$ wheels are governed by the Ackermann steering geometry, considering the wheelbase $l$ and track width $w$ of the vehicle:
$
\left\{
\begin{matrix} 
\delta_l = \textup{tan}^{-1}\left(\frac{2*l*\textup{tan}(\delta)}{2*l+w*\textup{tan}(\delta)}\right) \\ 
\delta_r = \textup{tan}^{-1}\left(\frac{2*l*\textup{tan}(\delta)}{2*l-w*\textup{tan}(\delta)}\right) 
\end{matrix}
\right.
$.

\paragraph{Suspension Dynamics}
\label{Suspension Dynamics}

For small and mid-scale vehicles, the suspension force acting on each sprung mass is calculated as ${^iM} * {^i{\ddot{Z}}} + {^iB} * ({^i{\dot{Z}}}-{^i{\dot{z}}}) + {^iK} * ({^i{Z}}-{^i{z}})$, where $^iZ$ and $^iz$ denote the displacements of the sprung and unsprung masses, respectively, and $^iB$ and $^iK$ represent the damping and spring coefficients of the $i$-th suspension. For full-scale vehicles, however, the stiffness ${^iK} = {^iM} * {^i\omega_n}^2$ and damping $^iB = 2 * ^i\zeta * \sqrt{{^iK} * {^iM}}$ coefficients of the suspension system are computed based on the sprung mass ${^iM}$, natural frequency ${^i\omega_n}$, and damping ratio ${^i\zeta}$ parameters. The point of suspension force application ${^iZ_F}$ is calculated based on the suspension geometry:
${^iZ_F} = {^iZ_\text{COM}} - {^iZ_w} + {^ir_w} - {^iZ_f}$, where $^iZ_\text{COM}$ denotes the Z-component of vehicle's center of mass, $^iZ_w$ is the Z-component of the relative transformation between each wheel and the vehicle frame ($^VT_{w_i}$), $^ir_w$ is the wheel radius, and $^iZ_f$ is the force offset determined by the suspension geometry. Lastly, the suspension displacement $^iZ_s$ at any given moment can be computed as ${^iZ_s} = \frac{{^iM} * g}{{^iZ_0} * {^iK}}$, where $g$ represents the acceleration due to gravity, and $^iZ_0$ is the suspension's equilibrium point. Additionally, full-scale vehicle models also have a provision to include anti-roll bars, which apply a force on the left ${^LF_r} = K_r * {^RZ} - {^LZ}$ and right ${R^F_r} = K_r * {^LZ} - {^RZ}$ wheels as long as they are grounded at the contact point $Z_c$. This force is directly proportional to the stiffness of the anti-roll bar, $K_r$. The left and right wheel travels are given by ${^LZ} = \frac{-{^LZ_c} - {^Lr_w}}{^LZ_s}$ and ${^RZ} = \frac{-{^RZ_c} - {^Rr_w}}{^RZ_s}$.

\paragraph{Tire Dynamics}
\label{Tire Dynamics}

Tire forces are determined based on the friction curve for each tire $\left\{\begin{matrix} {^iF_{t_x}} = F(^iS_x) \\{^iF_{t_y}} = F(^iS_y) \\ \end{matrix}\right.$, where $^iS_x$ and $^iS_y$ represent the longitudinal and lateral slips of the $i$-th tire, respectively. The friction curve is approximated using a two-piece spline, defined as $F(S) = \left\{\begin{matrix} f_0(S); \;\; S_0 \leq S < S_e \\ f_1(S); \;\; S_e \leq S < S_a \\ \end{matrix}\right.$, with $f_k(S) = a_k*S^3+b_k*S^2+c_k*S+d_k$ as a cubic polynomial function. The first segment of the spline ranges from zero $(S_0,F_0)$ to an extremum point $(S_e,F_e)$, while the second segment ranges from the extremum point $(S_e, F_e)$ to an asymptote point $(S_a, F_a)$. Tire slip is influenced by factors including tire stiffness $^iC_\alpha$, steering angle $\delta$, wheel speeds $^i\omega$, suspension forces $^iF_s$, and rigid-body momentum ${^iP}={^iM}*{^iv}$. The longitudinal slip $^iS_x$ of $i$-th tire is calculated by comparing the longitudinal components of its surface velocity $v_x$ (i.e., the longitudinal linear velocity of the vehicle) with its angular velocity $^i\omega$: ${^iS_x} = \frac{{^ir}*{^i\omega}-v_x}{v_x}$. The lateral slip $^iS_y$ depends on the tire's slip angle $\alpha$ and is determined by comparing the longitudinal $v_x$ (forward velocity) and lateral $v_y$ (side-slip velocity) components of the vehicle's linear velocity: ${^iS_y} = \tan(\alpha) = \frac{v_y}{\left| v_x \right|}$.

\paragraph{Aerodynamics}
\label{Aerodynamics}

Small and mid-scale vehicles are modeled with constant coefficients for linear $F_d$ as well as angular $T_d$ drags, which act directly proportional to their linear $v$ and angular $\omega$ velocities. These vehicles do not create significant downforce due to unoptimized aerodynamics, limited velocities and smaller size and mass. Full-scale vehicles, on the other hand, have been modeled to simulate variable air drag $F_\text{aero}$ acting on the vehicle, which is computed based on the vehicle’s operating condition:
$
F_{\text{aero}} = \begin{cases}
F_{d_\text{max}} & \text{if } v \geq v_{\text{max}} \\
F_{d_\text{idle}} & \text{if } \tau_{\text{out}} = 0 \\
F_{d_\text{rev}} & \text{if } (v \geq v_{\text{rev}}) \land (G_\# = -1) \land (RPM_{w} < 0) \\
F_{d_\text{idle}} & \text{otherwise}
\end{cases}
$
where, $v$ is the vehicle velocity, $v_\text{max}$ is the vehicle's designated top-speed, $v_\text{rev}$ is the vehicle's designated maximum reverse velocity, $G_\#$ is the operating gear, and $RPM_w$ is the average wheel RPM. The downforce acting on a full-scale vehicle is modeled proportional to its velocity: $F_\text{down}=K_\text{down}*|v|$, where $K_\text{down}$ is the downforce coefficient.

\subsubsection{Sensor Models}
\label{Sensor Models}

The simulated vehicles can be equipped with physically accurate interoceptive and exteroceptive sensing modalities.

\paragraph{Actuator Feedbacks}
\label{Actuator Feedbacks}

Throttle ($\tau$) and steering ($\delta$) sensors are simulated using a simple feedback loop.

\paragraph{Incremental Encoders}
\label{Incremental Encoders}

Simulated incremental encoders measure wheel rotations $^iN_{\text{ticks}} = {^iPPR} \times {^iCGR} \times {^iN_{\text{rev}}}$, where $^iN_{\text{ticks}}$ represents the measured ticks, $^iPPR$ is the encoder resolution (pulses per revolution), $^iCGR$ is the cumulative gear ratio, and $^iN_{\text{rev}}$ represents the wheel revolutions.

\paragraph{Inertial Navigation Systems}
\label{Inertial Navigation Systems}

Positioning systems and inertial measurement units (IMU) are simulated based on temporally coherent rigid-body transform updates of the vehicle $\{v\}$ with respect to the world $\{w\}$: ${^w\mathbf{T}_v} = \left[\begin{array}{c | c} \mathbf{R}_{3 \times 3} & \mathbf{t}_{3 \times 1} \\ \hline \mathbf{0}_{1 \times 3} & 1 \end{array}\right] \in SE(3)$. The positioning systems provide 3-DOF positional coordinates $\{x,y,z\}$ of the vehicle, while the IMU supplies linear accelerations $\{a_x,a_y,a_z\}$, angular velocities $\{\omega_x,\omega_y,\omega_z\}$, and 3-DOF orientation data for the vehicle, either as Euler angles $\{\phi_x,\theta_y,\psi_z\}$ or as a quaternion $\{q_0,q_1,q_2,q_3\}$.

\paragraph{Planar LIDARs}
\label{Planar LIDARs}

2D LIDAR simulation employs iterative ray-casting \texttt{raycast}\{$^w\mathbf{T}_l$, $\vec{\mathbf{R}}$, $r_{\text{max}}$\} for each angle $\theta \in \left [ \theta_{\text{min}}:\theta_{\text{res}}:\theta_{\text{max}} \right ]$ at a specified update rate. Here, ${^w\mathbf{T}_l} = {^w\mathbf{T}_v} * {^v\mathbf{T}_l} \in SE(3)$ represents the relative transformation of the LIDAR \{$l$\} with respect to the vehicle \{$v$\} and the world \{$w$\}, $\vec{\mathbf{R}} = \left [\cos(\theta) \;\; \sin(\theta) \;\; 0 \right ]^T$ defines the direction vector of each ray-cast $R$, where $r_{\text{min}}$ and $r_{\text{max}}$ denote the minimum and maximum linear ranges, $\theta_{\text{min}}$ and $\theta_{\text{max}}$ denote the minimum and maximum angular ranges, and $\theta_{\text{res}}$ represents the angular resolution of the LIDAR, respectively. The laser scan ranges are determined by checking ray-cast hits and then applying a threshold to the minimum linear range of the LIDAR, calculated as \texttt{ranges[i]}$=\begin{cases} \texttt{hit.dist} & \text{ if } \texttt{ray[i].hit} \text{ and } \texttt{hit.dist} \geq r_{\text{min}} \\ \infty & \text{ otherwise} \end{cases}$, where \texttt{ray.hit} is a Boolean flag indicating whether a ray-cast hits any colliders in the scene, and \texttt{hit.dist}$=\sqrt{(x_{\text{hit}}-x_{\text{ray}})^2 + (y_{\text{hit}}-y_{\text{ray}})^2 + (z_{\text{hit}}-z_{\text{ray}})^2}$ calculates the Euclidean distance from the ray-cast source $\{x_{\text{ray}}, y_{\text{ray}}, z_{\text{ray}}\}$ to the hit point $\{x_{\text{hit}}, y_{\text{hit}}, z_{\text{hit}}\}$.

\paragraph{Spatial LIDARs}
\label{Spatial LIDARs}

3D LIDAR simulation adopts multi-channel parallel ray-casting \texttt{raycast}\{$^w\mathbf{T}_l$, $\vec{\mathbf{R}}$, $r_{\text{max}}$\} for each angle $\theta \in \left [ \theta_{\text{min}}:\theta_{\text{res}}:\theta_{\text{max}} \right ]$ and each channel $\phi \in \left [ \phi_{\text{min}}:\phi_{\text{res}}:\phi_{\text{max}} \right ]$ at a specified update rate, with GPU acceleration (if available). Here, ${^w\mathbf{T}_l} = {^w\mathbf{T}_v} * {^v\mathbf{T}_l} \in SE(3)$ represents the relative transformation of the LIDAR \{$l$\} with respect to the vehicle \{$v$\} and the world \{$w$\}, $\vec{\mathbf{R}} = \left [\cos(\theta)*\cos(\phi) \;\; \sin(\theta)*\cos(\phi) \;\; -\sin(\phi) \right ]^T$ defines the direction vector of each ray-cast $R$, where $r_{\text{min}}$ and $r_{\text{max}}$ denote the minimum and maximum linear ranges, $\theta_{\text{min}}$ and $\theta_{\text{max}}$ denote the minimum and maximum horizontal angular ranges, $\phi_{\text{min}}$ and $\phi_{\text{max}}$ denote the minimum and maximum vertical angular ranges, and $\theta_{\text{res}}$ and $\phi_{\text{res}}$ represent the horizontal and vertical angular resolutions of the LIDAR, respectively. The thresholded ray-cast hit coordinates $\{x_{\text{hit}}, y_{\text{hit}}, z_{\text{hit}}\}$, from each of the casted rays is encoded into byte arrays based on the LIDAR parameters, and given out as the point cloud data.

\paragraph{Cameras}
\label{Cameras}

Simulated cameras are parameterized by their focal length $f=$, sensor size $\{s_x, s_y\}$, target resolution, as well as the distances to the near $N$ and far $F$ clipping planes. The viewport rendering pipeline for the simulated cameras operates in three stages. First, the camera view matrix $\mathbf{V} \in SE(3)$ is computed by obtaining the relative homogeneous transform of the camera $\{c\}$ with respect to the world $\{w\}$: $\mathbf{V} = \begin{bmatrix} r_{00} & r_{01} & r_{02} & t_{0} \\ r_{10} & r_{11} & r_{12} & t_{1} \\ r_{20} & r_{21} & r_{22} & t_{2} \\ 0 & 0 & 0 & 1 \\ \end{bmatrix}$, where $r_{ij}$ and $t_i$ denote the rotational and translational components, respectively. Next, the camera projection matrix $\mathbf{P} \in \mathbb{R}^{4 \times 4}$ is calculated to project world coordinates into image space coordinates: $\mathbf{P} = \begin{bmatrix} \frac{2*N}{R-L} & 0 & \frac{R+L}{R-L} & 0 \\ 0 & \frac{2*N}{T-B} & \frac{T+B}{T-B} & 0 \\ 0 & 0 & -\frac{F+N}{F-N} & -\frac{2*F*N}{F-N} \\ 0 & 0 & -1 & 0 \\ \end{bmatrix}$, where $L$, $R$, $T$, and $B$ denote the left, right, top, and bottom offsets of the sensor. The camera parameters $\{f,s_x,s_y\}$ are related to the terms of the projection matrix as follows: $f = \frac{2*N}{R-L}$, $a = \frac{s_y}{s_x}$, and $\frac{f}{a} = \frac{2*N}{T-B}$. The perspective projection from the simulated camera's viewport is given as $\mathbf{C} = \mathbf{P}*\mathbf{V}*\mathbf{W}$, where $\mathbf{C} = \left [x_c\;\;y_c\;\;z_c\;\;w_c \right ]^T$ represents image space coordinates, and $\mathbf{W} = \left [x_w\;\;y_w\;\;z_w\;\;w_w \right ]^T$ represents world coordinates. Finally, this camera projection is transformed into normalized device coordinates (NDC) by performing perspective division (i.e., dividing throughout by $w_c$), leading to a viewport projection achieved by scaling and shifting the result and then utilizing the rasterization process of the graphics API (e.g., DirectX for Windows, Metal for macOS, and Vulkan for Linux). Additionally, a post-processing step simulates non-linear lens and film effects, such as lens distortion, depth of field, exposure, ambient occlusion, contact shadows, bloom, motion blur, film grain, chromatic aberration, etc.

\subsubsection{Calibration and Validation}
\label{Sub-Section: Calibration and Validation}

The vehicle digital twin models were calibrated and validated against geometric, static and dynamic measurement data collected from their real-world counterparts and/or their datasheets. This included the validation of geometric measurements for physical as well as visual purposes, static calibration for mass and center of mass parameters and dynamic calibration for validating standard benchmark maneuvers performed in open-loop tests. Additionally, sensor models were validated against static and dynamic characteristics of their real-world counterparts. Fig. \ref{fig: figure6} depicts some of these calibration/validation tests.

\begin{figure}[t]
    \centering
    \includegraphics[width=\linewidth]{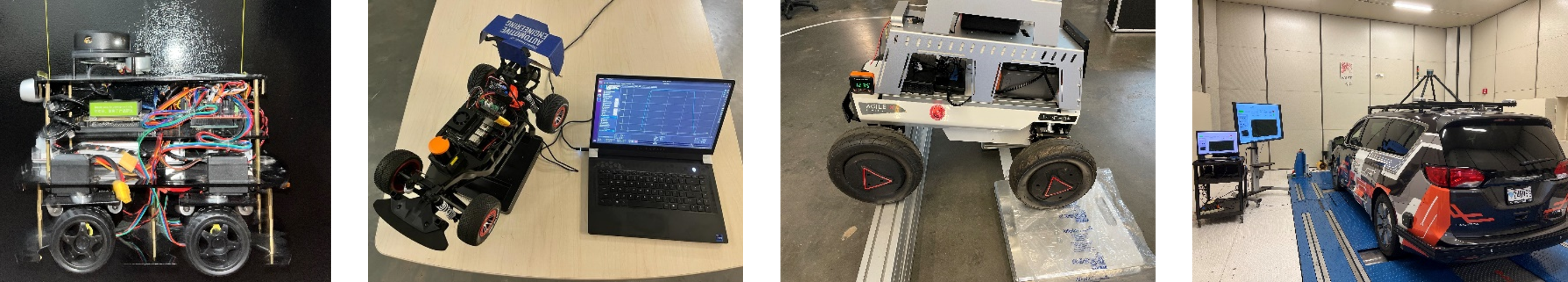}
    \caption{Calibration and validation of vehicle digital twins: CG position calibration/validation of Nigel, VESC calibration/validation of F1TENTH, static measurements of Hunter SE, and powertrain measurements of OpenCAV.}
    \label{fig: figure6}
\end{figure}

\subsubsection{Physical Vehicle Build and Setup}
\label{Sub-Section: Physical Vehicle Build and Setup}

We predominantly worked with two small-scale physical vehicles during the course of this project. These included an F1TENTH and a Nigel. While the prior was acquired off-the-shelf, its hardware and software had to be tested, configured, calibrated and set up to interface with the \href{https://github.com/Tinker-Twins/AutoDRIVE-F1TENTH}{ROS}, \href{https://github.com/Tinker-Twins/AutoDRIVE-F1TENTH}{ROS 2} and \href{https://github.com/Tinker-Twins/AutoDRIVE-Autoware}{Autoware} stacks. The latter, on the other hand, was designed, manufactured and assembled from scratch, including its \href{https://github.com/AutoDRIVE-Ecosystem/Nigel-SolidWorks}{mechanical} and \href{https://github.com/AutoDRIVE-Ecosystem/Nigel-Fritzing}{electronic} subsystems, \href{https://github.com/AutoDRIVE-Ecosystem/Nigel-Arduino}{firmware development} as well as \href{https://github.com/Tinker-Twins/AutoDRIVE/tree/AutoDRIVE-Devkit/ADSS Toolkit/autodrive_ros/autodrive_nigel}{ROS}, \href{https://github.com/Tinker-Twins/AutoDRIVE/tree/AutoDRIVE-Devkit/ADSS Toolkit/autodrive_ros2/autodrive_nigel}{ROS 2} and \href{https://github.com/Tinker-Twins/AutoDRIVE-Autoware}{Autoware} integration.

\begin{figure}[h]
    \centering
    \includegraphics[width=\linewidth]{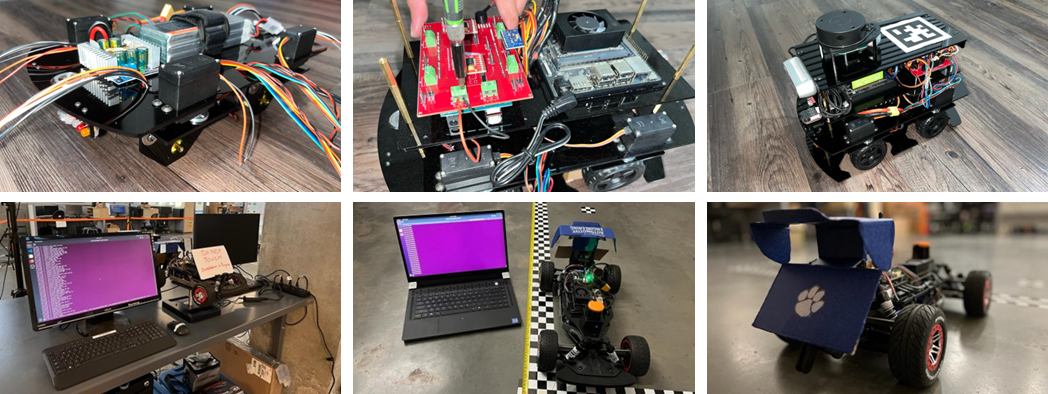}
    \caption{Physical build and setup depicting the progress stages during the hardware/software build, calibration and testing of F1TENTH and Nigel vehicles.}
    \label{fig: figure7}
\end{figure}

Being open-source vehicles, build documentation for \href{https://f1tenth.org/build.html}{F1TENTH} and \href{https://github.com/Tinker-Twins/AutoDRIVE/blob/AutoDRIVE-Testbed/Documents/Nigel - Assembly Guide.pdf}{Nigel} are readily available online. Fig. \ref{fig: figure7} depicts intermittent stages in building and setting up these vehicles.

\hypertarget{Environment Digital Twins}{%
\subsection{Environment Digital Twins}\label{Environment Digital Twins}}

\begin{figure}[h]
    \centering
    \includegraphics[width=\linewidth]{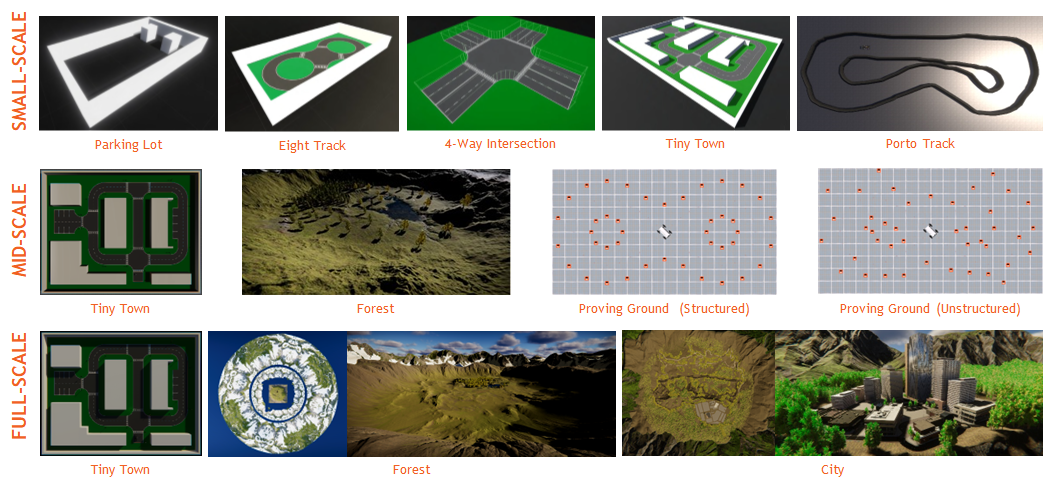}
    \caption{Autonomy-oriented environment digital twins across scales: Parking Lot, Eight Track, 4-Way Intersection, Tiny Town and Porto Track (small-scale), Tiny Town, Forest and Proving Ground (mid-scale), and Tiny Town, Forest and City (full-scale) scenarios for on/off-road autonomy.}
    \label{fig: figure8}
\end{figure}

As described earlier, we leveraged AutoDRIVE Simulator \cite{AutoDRIVESimulator, AutoDRIVESimulatorReport} to develop digital twin models of various environments, across different scales and ODDs (Fig. \ref{fig: figure8}). These included realistic counterparts of small-scale environments such as the Parking Lot, Eight Track, 4-Way Intersection and Tiny Town for Nigel, which were developed using AutoDRIVE IDK, as well as the Porto Track for F1TENTH, which was created based on the binary occupancy grid map of its real-world counterpart. Additionally, simplistic mid-scale and full-scale environments such as the scaled-up versions of Tiny Town along with structured and unstructured Proving Ground scenarios were developed. Finally, two highly detailed, albeit imaginary, mid-scale and full-scale scenarios were developed to support on-road as well as off-road autonomy. These included a City scenario and a Forest environment (Fig. \ref{figure9}). The full-scale variants of these scenarios have several rich features and are large enough to support driving for several minutes, if not a few hours. Additionally, these scenarios allowed the simulation of various environmental conditions, such as different times of day as well as weather conditions, to introduce additional degrees of variability. Finally, environmental physics was simulated accurately by conducting mesh-mesh interference detection and computing contact forces, frictional forces, momentum transfer, as well as linear and angular drag acting on all rigid bodies, at each time step.

\begin{figure}[h]
     \centering
     \begin{subfigure}[b]{0.49\linewidth}
         \centering
         \includegraphics[width=\linewidth]{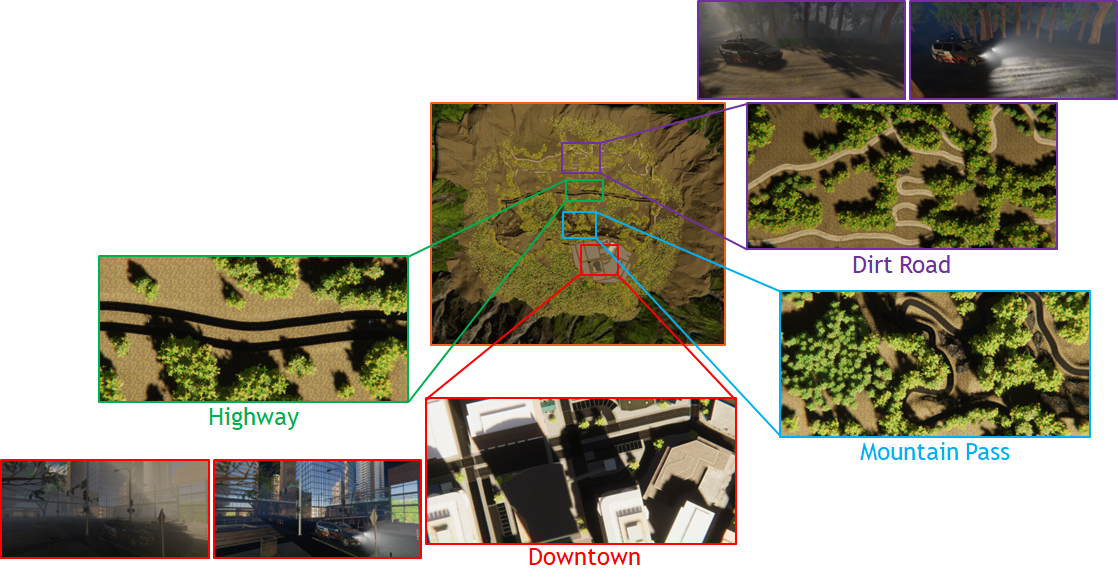}
         \caption{City}
         \label{fig9a}
     \end{subfigure}
     \hfill
     \begin{subfigure}[b]{0.49\linewidth}
         \centering
         \includegraphics[width=\linewidth]{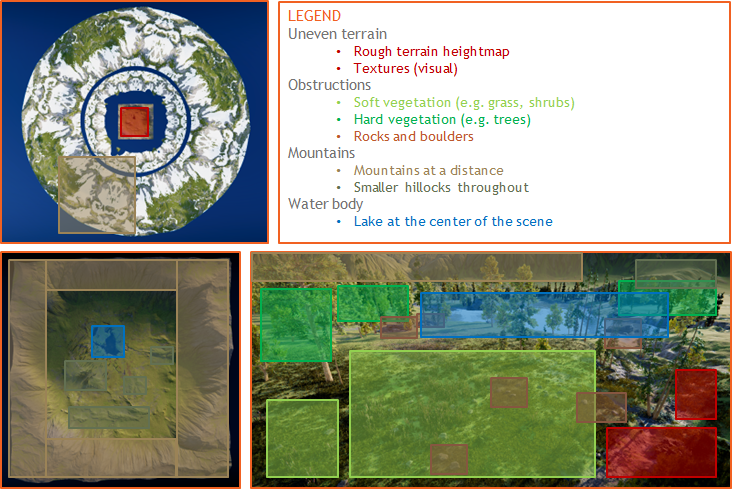}
         \caption{Forest}
         \label{fig9b}
     \end{subfigure}
     \caption{City and Forest environment digital twins depicting their respective highlights and features to support on-road as well as off-road autonomy.}
    \label{figure9}
\end{figure}

From a computational perspective, the said environment digital twins, especially the full-scale scenarios, made use of pre-baked lightmaps, which provided the benefits of physics-based lighting while reducing the computational overhead of real-time raytracing. Additionally, the simulator implemented level-of-detail (LOD) culling to gradually degrade the LOD of environmental objects as they moved further away from the scene cameras. However, it was ensured that LOD culling did not affect any of the AV camera sensor(s).

AutoDRIVE Simulator supports various approaches to developing realistic/imaginary environment digital twins:
\begin{itemize}
     \item \textit{AutoDRIVE IDK:} Custom scenarios and maps can be crafted by utilizing the modular and adaptable Infrastructure Development Kit (IDK). This kit provides the flexibility to configure terrain modules, road networks, obstruction modules, and traffic elements. Specifically, the Parking Lot, Eight Track, 4-Way Intersection, Tiny Town and Proving Ground scenarios were developed using AutoDRIVE IDK.

     \item \textit{Plug-In Scenarios:} AutoDRIVE Simulator supports third-party tools, such as RoadRunner \cite{RoadRunner}, and open standards like OpenSCENARIO \cite{OpenSCENARIO} and OpenDRIVE \cite{OpenDRIVE}). This allows users to incorporate a diverse range of plugins, packages, and assets in several standard formats for creating or customizing driving scenarios. Particularly, the autonomous racing scenario was created based on the binary occupancy grid map of a real-world F1TENTH racetrack called ``Porto'' using a third-party 3D modeling software, which was then imported into AutoDRIVE Simulator and post-processed with physical as well as graphical enhancements to make it ``sim-ready''.

     \item \textit{Unity Terrain Integration:} Since the AutoDRIVE Simulator is built atop the Unity \cite{Unity} game engine, it seamlessly supports scenario design and development through Unity Terrain \cite{UnityTerrain}. Users have the option to define terrain meshes, textures, heightmaps, vegetation, skyboxes, wind effects, and more, allowing the design of both on-road and off-road scenarios. This option is well-suited for modeling full-scale environments. Specifically, Forest and City scenes were developed using this technique.
\end{itemize}

\hypertarget{APIs and HMIs to Connect with Digital Twins}{%
\subsection{APIs and HMIs to Connect with Digital Twins}\label{APIs and HMIs to Connect with Digital Twins}}

\begin{figure}[h]
    \centering
    \includegraphics[width=\linewidth]{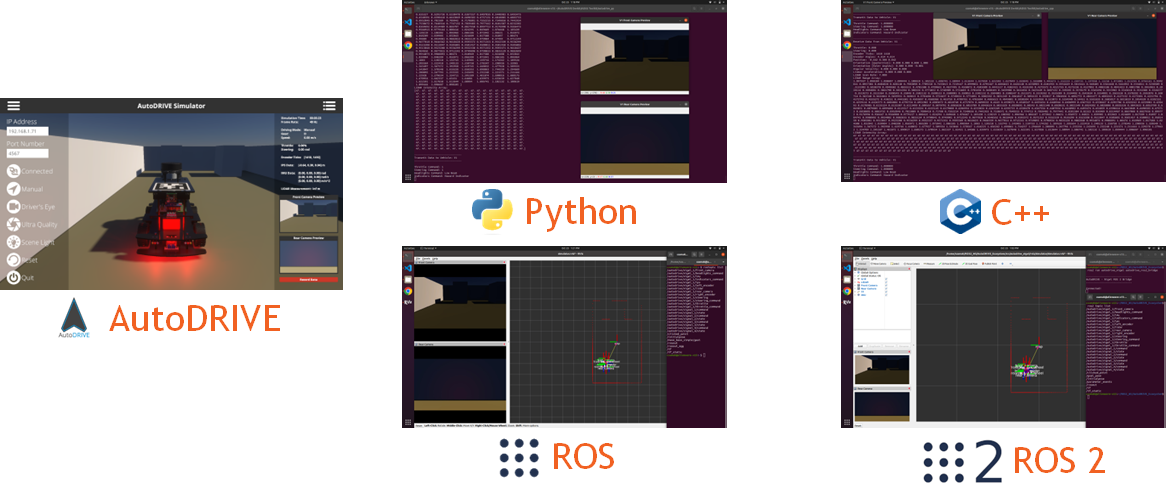}
    \caption{APIs to connect with AutoDRIVE Ecosystem: Python, C++, ROS and ROS 2. Note that AutoDRIVE-Autoware integration is accomplished by extending the ROS 2 API.}
    \label{fig: figure10}
\end{figure}

The integration of APIs within the AutoDRIVE Ecosystem was achieved through the comprehensive expansion and incorporation of AutoDRIVE Devkit. The versatile APIs developed as part of this framework facilitate interactions with the virtual/real vehicles using Python, C++, ROS, ROS 2, or the Autoware stack (Fig. \ref{fig: figure10}). This expansion caters to a diverse range of programming preferences, empowering users to exploit the AutoDRIVE Simulator or AutoDRIVE Testbed for swift and flexible development of autonomy algorithms. The framework extends its utility by enabling the development of API-mediated HMIs, catering to both virtual and physical vehicles.

\begin{figure}[t]
    \centering
    \includegraphics[width=\linewidth]{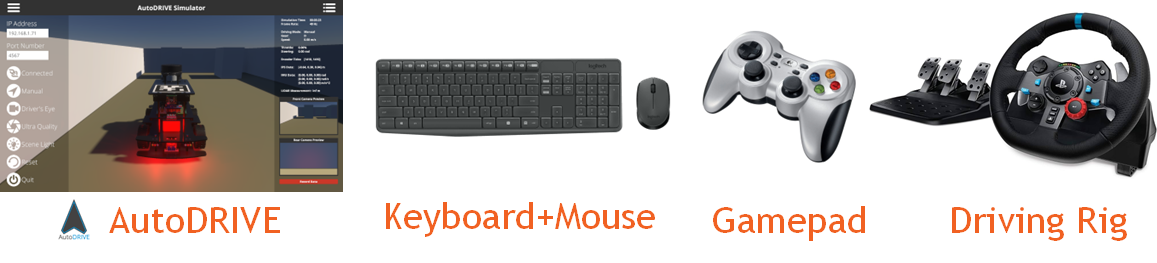}
    \caption{HMIs to connect with AutoDRIVE Ecosystem include standard keyboard (digital) and mouse (analog), gamepad/joystick (analog) as well as driving and steering rigs (hybrid).}
    \label{fig: figure11}
\end{figure}

Apart from the API-mediated HMIs, the simulation framework itself served a dual purpose by not only providing a digital twinning platform, but also enabling the development of direct HMIs to interface with the virtual vehicles (Fig. \ref{fig: figure11}). This direct-HMI teleoperation framework, designed for scalability, ensures practical feasibility by relaying identical machine-to-machine (M2M) commands to both virtual and real vehicles. The versatility of this approach allows for a true digital-twin framework, establishing a seamless connection between the simulated environment and the physical world. Additionally, in an extended-reality (XR) setup, this framework offers opportunities to extend the direct-HMI teleoperation to real vehicles, enhancing the applicability and potential of AutoDRIVE Ecosystem in diverse operational scenarios.
\pagebreak
\hypertarget{Autoware Deployment Results}{%
\section{Autoware Deployment Results}\label{Autoware Deployment Results}}

The two of us worked with the Autoware stack for the very first time while operating and experimenting with the OpenCAV at CU-ICAR. This full-scale modular, open-architecture, open-interface, and open-source-software-based research instrument gave us ample opportunity to look ``under the hood'' and understand how this complex system-of-systems functions autonomously.

Furthermore, the format of the operational demo of OpenCAV shuttle rides across CU-ICAR campus motivated us to define the exact problem statement for the Autoware-based autonomy deployment segment of the project. This end-to-end autonomy pipeline works in 3 stages.

\begin{enumerate}
    \item First, the environment is mapped using the LIDAR point cloud data and optionally using odometry estimates by fusing IMU and encoder data while driving (or teleoperating) the vehicle manually.
    \item Next, a reference trajectory is generated by manually driving the vehicle within the (pre)mapped environment, while recording the waypoint coordinates as the vehicle localization estimates space a certain threshold distance apart with respect to the map's coordinate frame (using the LIDAR point cloud data and optionally using odometry estimates by fusing IMU and encoder data). It is worth mentioning that a reference trajectory can also be defined completely offline by using the map information alone, however, in such a scenario, proper care needs to be taken to ensure that the resulting trajectory is completely safe and kinodynamically feasible for the vehicle to follow in autonomous mode.
    \item Finally, in autonomous mode, the vehicle tracks the reference trajectory using a linearized pure-pursuit controller for lateral motion control and PID controller for longitudinal motion control.
\end{enumerate}

The exact inputs, outputs, and configurations of perception, planning, and control modules vary with the underlying vehicle platform. Therefore, to keep the overall project clean and well-organized, a multitude of custom meta-packages were developed within the Autoware Universe stack to handle different perception, planning, and control algorithms using different input and output information in the form of independent individual packages.

Additionally, a separate meta-package was created to handle different vehicles of the AutoDRIVE Ecosystem exploited in this project viz. Nigel, F1TENTH, Hunter SE and OpenCAV. Each package for a particular vehicle hosts vehicle-specific parameter description configuration files for perception, planning, and control algorithms, map files, RViz configuration files, API program files, teleoperation program files, and user-convenient launch files for getting started quickly and easily.

It is worth mentioning that we have also implemented a couple of high-level system configurations within the Autoware stack to ensure proper operation of the autonomous vehicles.

\begin{enumerate}
    \item A trajectory looping criteria to determine if the very first waypoint in the reference trajectory is to be set as the target waypoint for the controller upon successful tracking of the very last waypoint in the reference trajectory. This helps effectively shape a ``safe termination'' behavior for the ego vehicle upon the end of the reference trajectory, which is extremely important in certain applications (e.g. autonomous valet parking) as discussed in further sections. Furthermore, the tolerance on the termination criteria helps the ego vehicle come to a safe stop smoothly without any unnecessarily aggressive over-corrections or any overshoots outside the safe space.
    \item An operational control mode for the vehicle to engage the ego vehicle selectively in a simplistic gym environment, high-fidelity simulation environment, purely testbed (real-world hardware) mode, or in a true digital twin framework exploiting the high-fidelity simulator in conjunction with the testbed.
\end{enumerate}

\hypertarget{F1TENTH}{%
\subsection{F1TENTH}\label{F1TENTH}}

\begin{figure}[H]
    \centering
    \includegraphics[width=\linewidth]{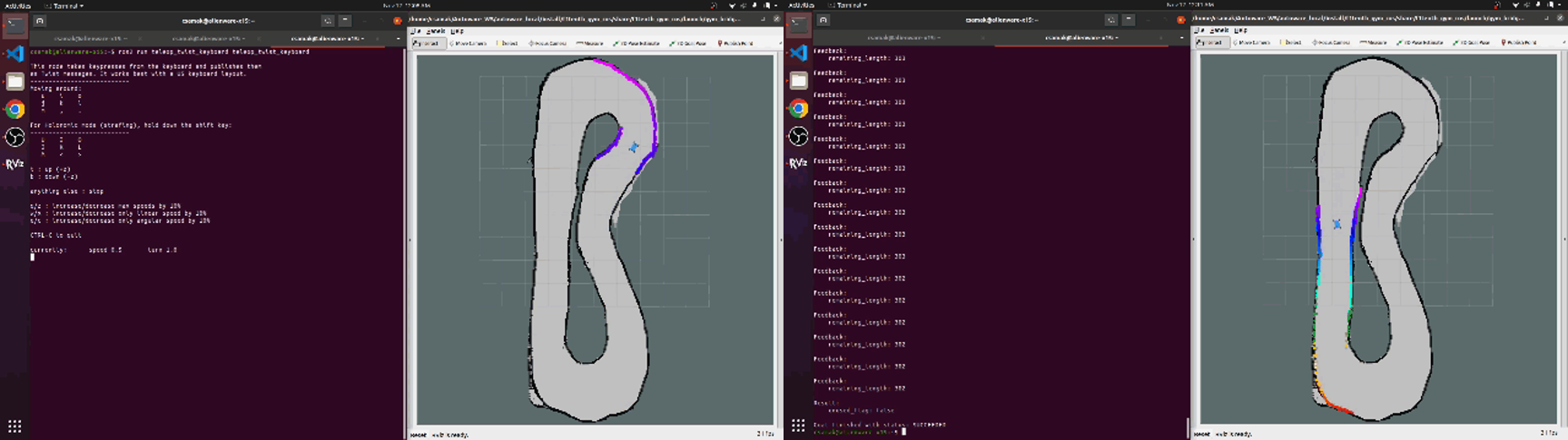}
    \caption{AutoDRIVE-F1TENTH-Autoware integration for autonomous racing ODD: RViz Gym demo for recording a trajectory using manual teleoperation and then tracking the pre-recorded trajectory autonomously.}
    \label{fig: figure12}
\end{figure}

\begin{figure}[H]
    \centering
    \includegraphics[width=\linewidth]{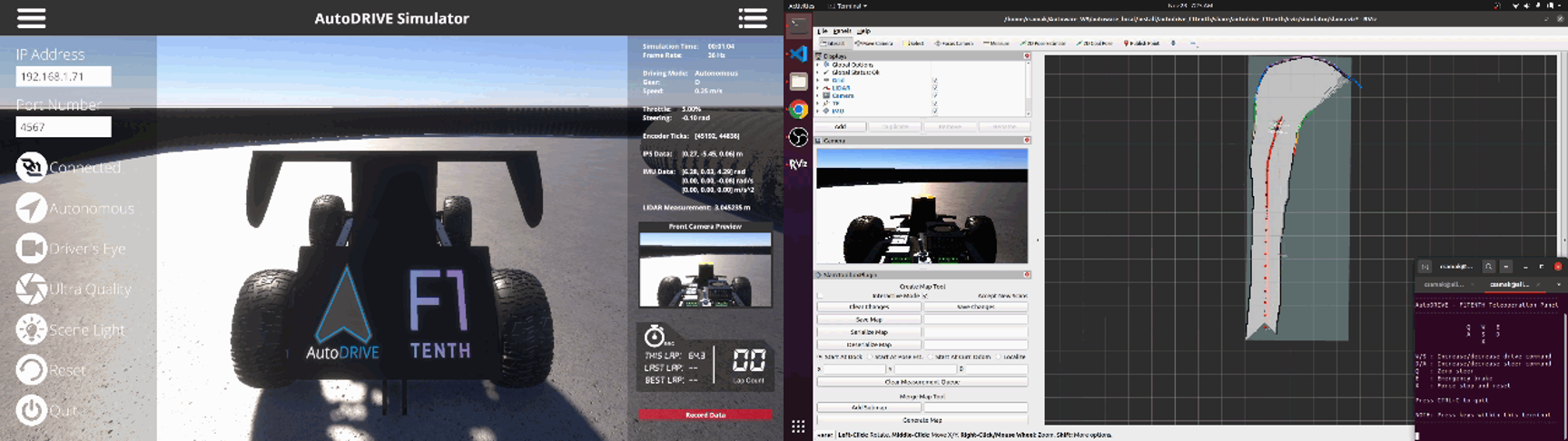}
    \caption{AutoDRIVE-F1TENTH-Autoware integration for autonomous racing ODD: AutoDRIVE Simulator demo for mapping an unknown racetrack using manual teleoperation.}
    \label{fig: figure13}
\end{figure}

\begin{figure}[H]
    \centering
    \includegraphics[width=\linewidth]{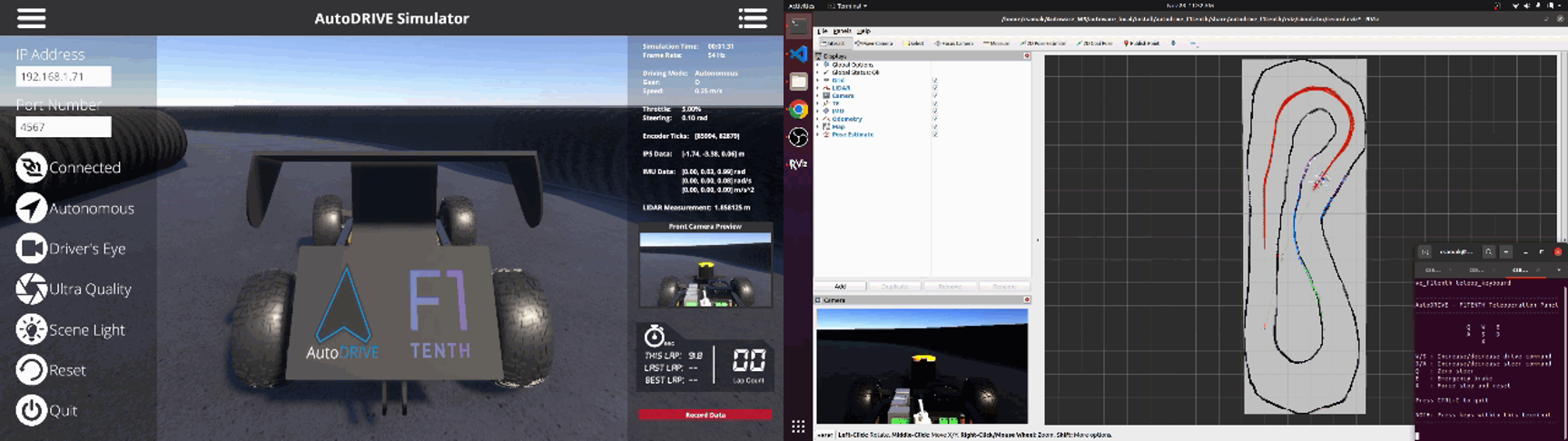}
    \caption{AutoDRIVE-F1TENTH-Autoware integration for autonomous racing ODD: AutoDRIVE Simulator demo for recording a trajectory using manual teleoperation.}
    \label{fig: figure14}
\end{figure}

\begin{figure}[H]
    \centering
    \includegraphics[width=\linewidth]{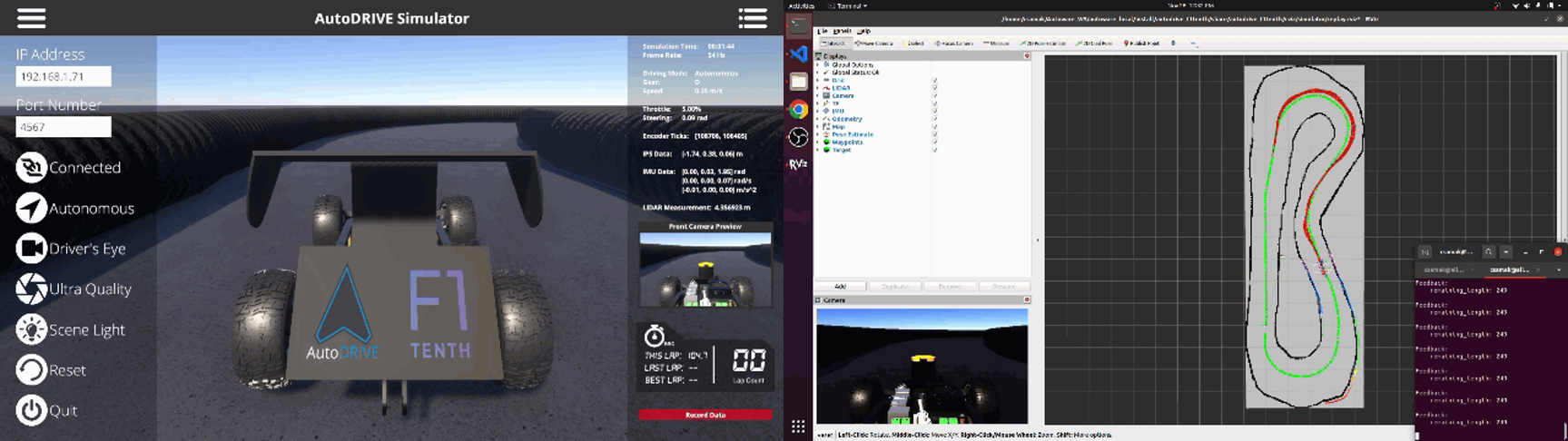}
    \caption{AutoDRIVE-F1TENTH-Autoware integration for autonomous racing ODD: AutoDRIVE Simulator demo for tracking the pre-recorded trajectory autonomously.}
    \label{fig: figure15}
\end{figure}

\begin{figure}[H]
    \centering
    \includegraphics[width=\linewidth]{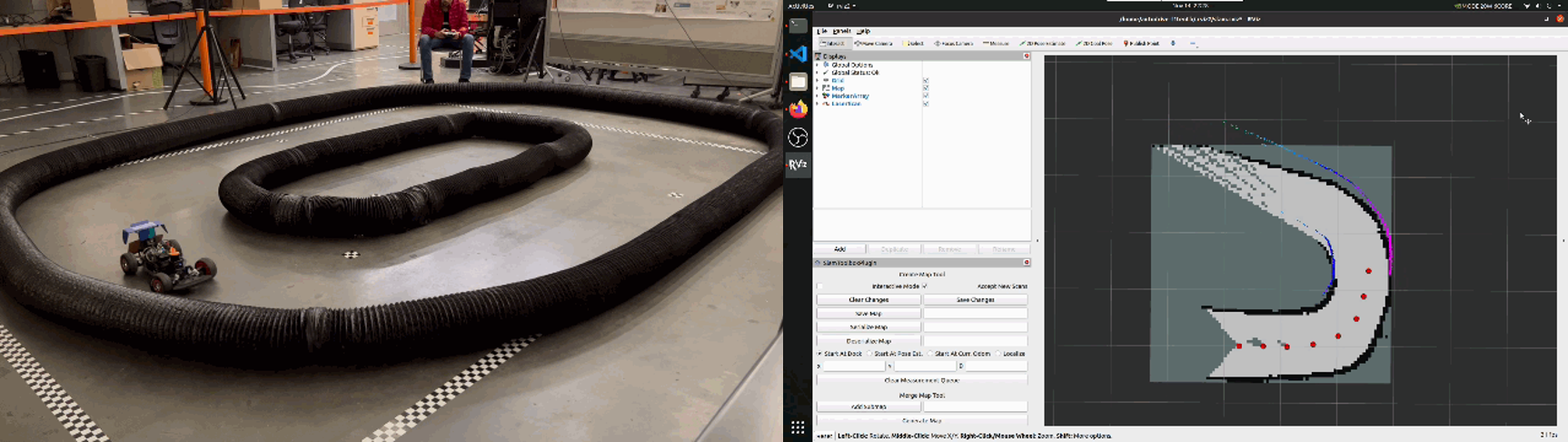}
    \caption{AutoDRIVE-F1TENTH-Autoware integration for autonomous racing ODD: Real-world demo for mapping an unknown racetrack using manual teleoperation.}
    \label{fig: figure16}
\end{figure}

\begin{figure}[H]
    \centering
    \includegraphics[width=\linewidth]{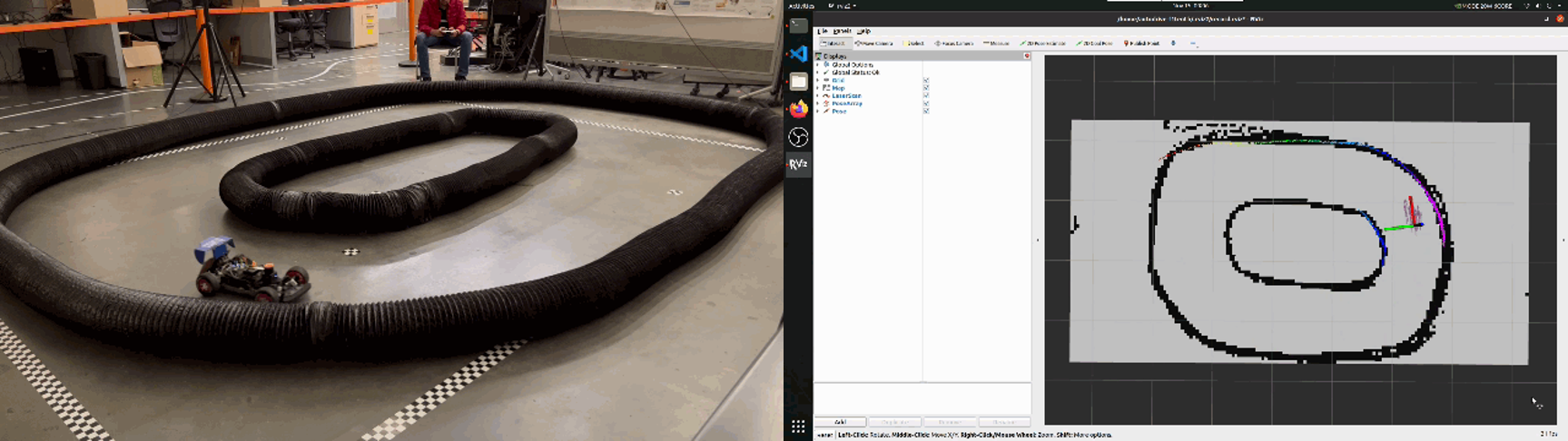}
    \caption{AutoDRIVE-F1TENTH-Autoware integration for autonomous racing ODD: Real-world demo for recording a trajectory using manual teleoperation.}
    \label{fig: figure17}
\end{figure}

\begin{figure}[H]
    \centering
    \includegraphics[width=\linewidth]{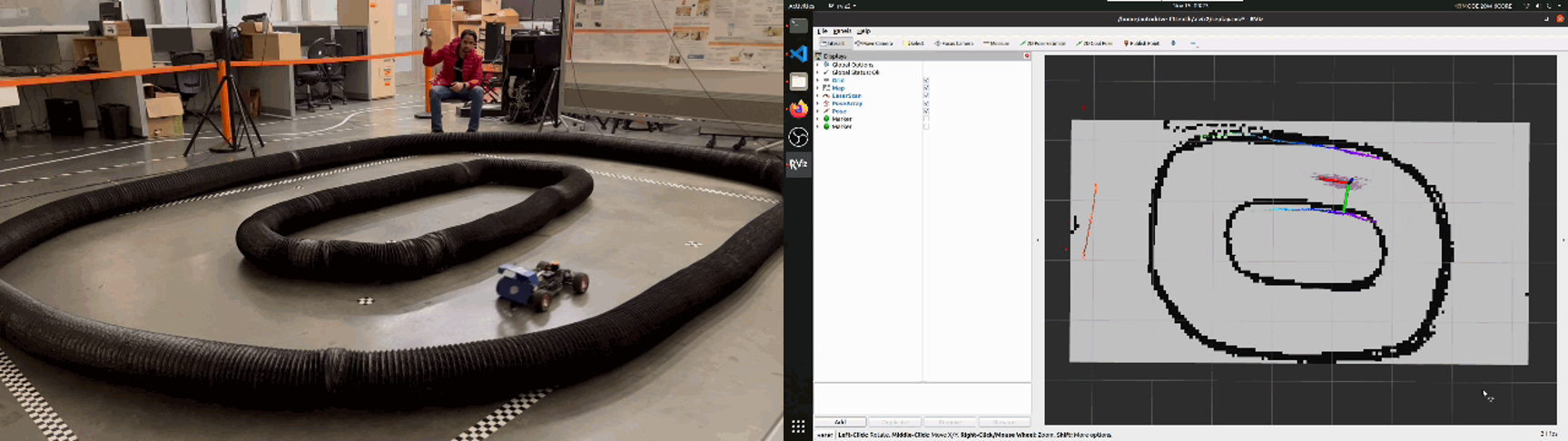}
    \caption{AutoDRIVE-F1TENTH-Autoware integration for autonomous racing ODD: Real-world demo for tracking the pre-recorded trajectory autonomously.}
    \label{fig: figure18}
\end{figure}

\hypertarget{Nigel}{%
\subsection{Nigel}\label{Nigel}}

\begin{figure}[H]
    \centering
    \includegraphics[width=\linewidth]{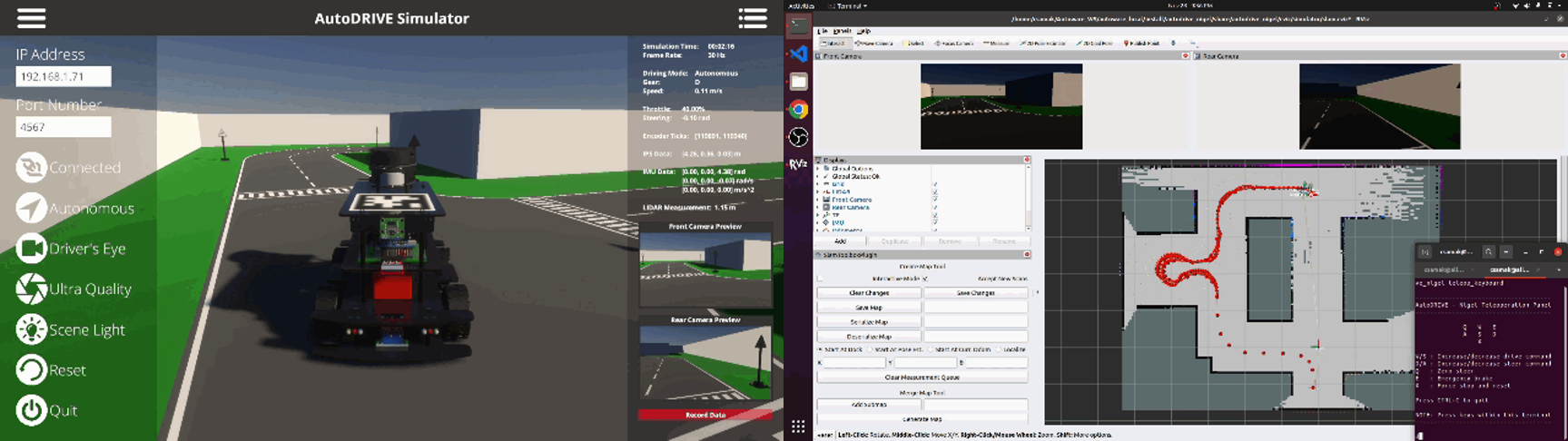}
    \caption{AutoDRIVE-Nigel-Autoware integration for autonomous valet parking ODD: AutoDRIVE Simulator demo for mapping an unknown environment using manual teleoperation.}
    \label{fig: figure19}
\end{figure}

\begin{figure}[H]
    \centering
    \includegraphics[width=\linewidth]{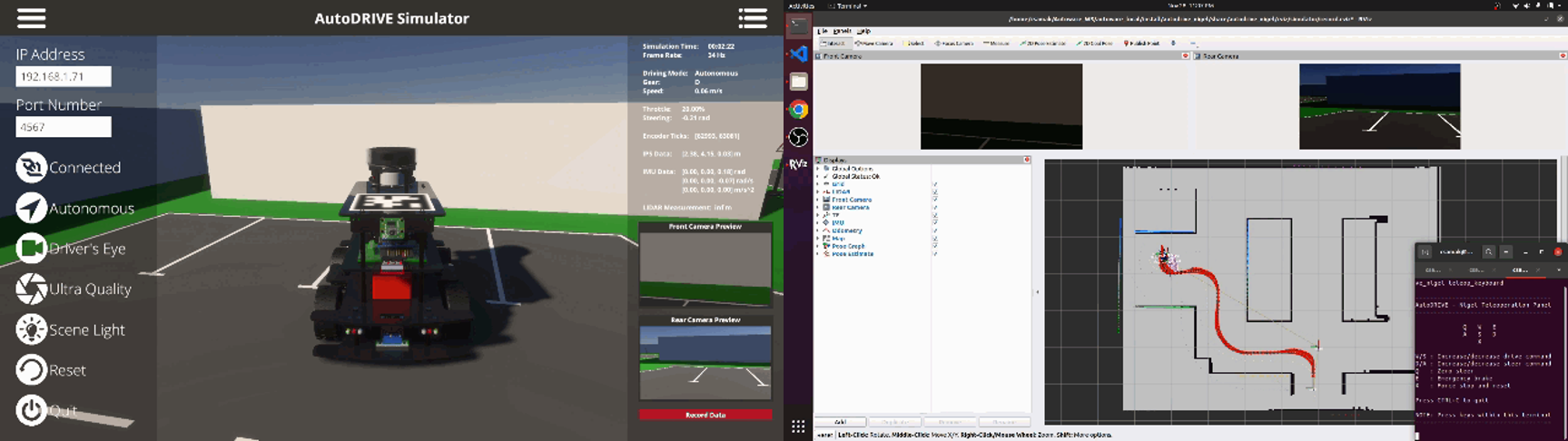}
    \caption{AutoDRIVE-Nigel-Autoware integration for autonomous valet parking ODD: AutoDRIVE Simulator demo for recording a trajectory using manual teleoperation.}
    \label{fig: figure20}
\end{figure}

\begin{figure}[H]
    \centering
    \includegraphics[width=\linewidth]{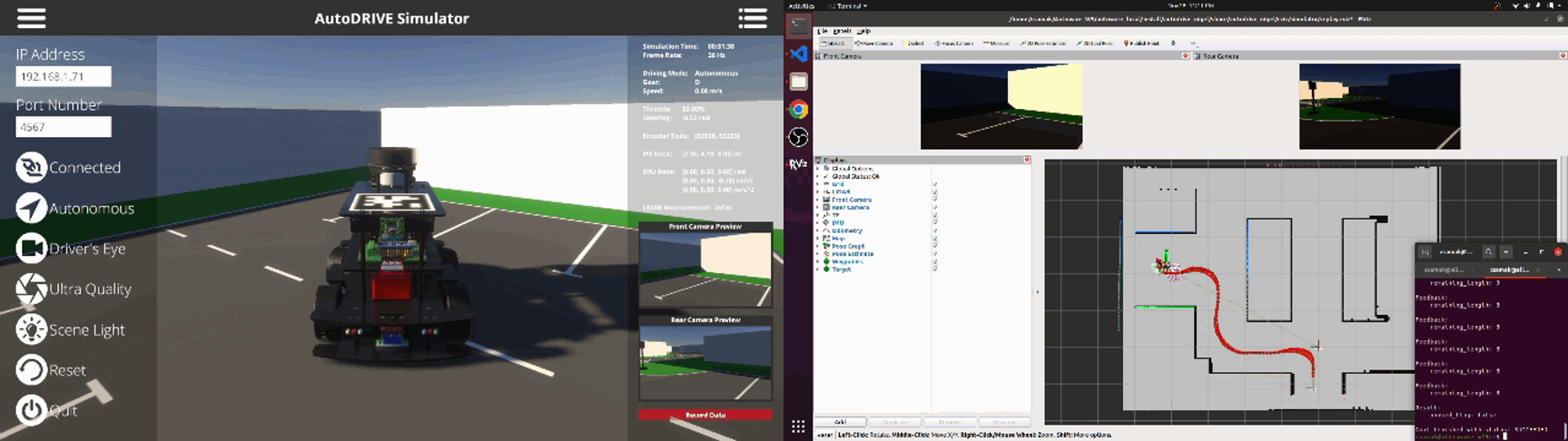}
    \caption{AutoDRIVE-Nigel-Autoware integration for autonomous valet parking ODD: AutoDRIVE Simulator demo for tracking the pre-recorded trajectory autonomously.}
    \label{fig: figure21}
\end{figure}

\begin{figure}[H]
    \centering
    \includegraphics[width=\linewidth]{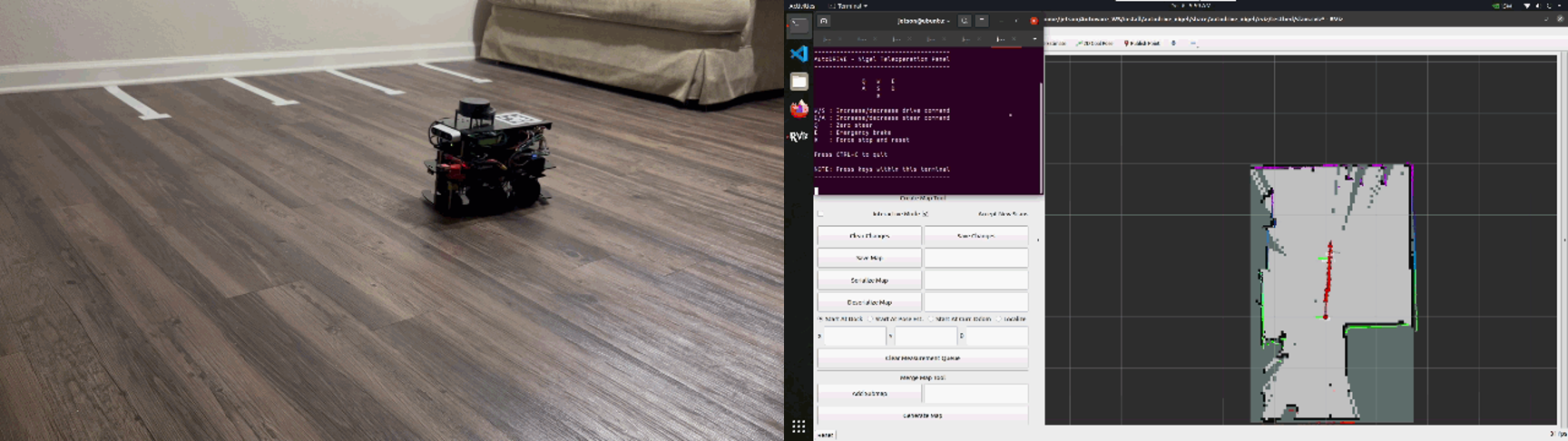}
    \caption{AutoDRIVE-Nigel-Autoware integration for autonomous valet parking ODD: Real-world demo for mapping an unknown environment using manual teleoperation.}
    \label{fig: figure22}
\end{figure}

\begin{figure}[H]
    \centering
    \includegraphics[width=\linewidth]{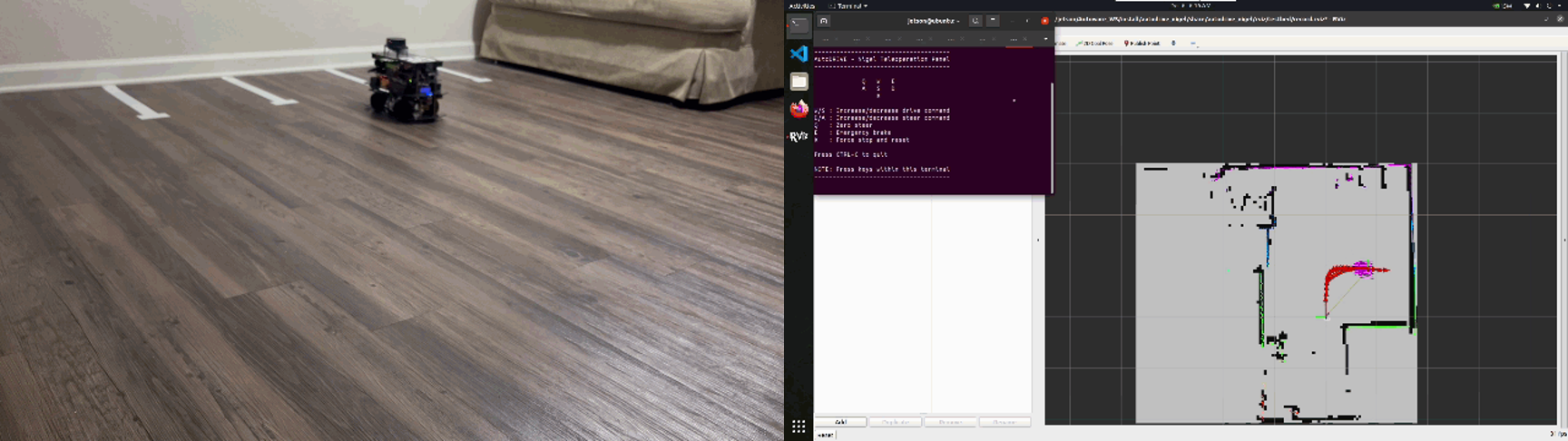}
    \caption{AutoDRIVE-Nigel-Autoware integration for autonomous valet parking ODD: Real-world demo for recording a trajectory using manual teleoperation.}
    \label{fig: figure23}
\end{figure}

\begin{figure}[H]
    \centering
    \includegraphics[width=\linewidth]{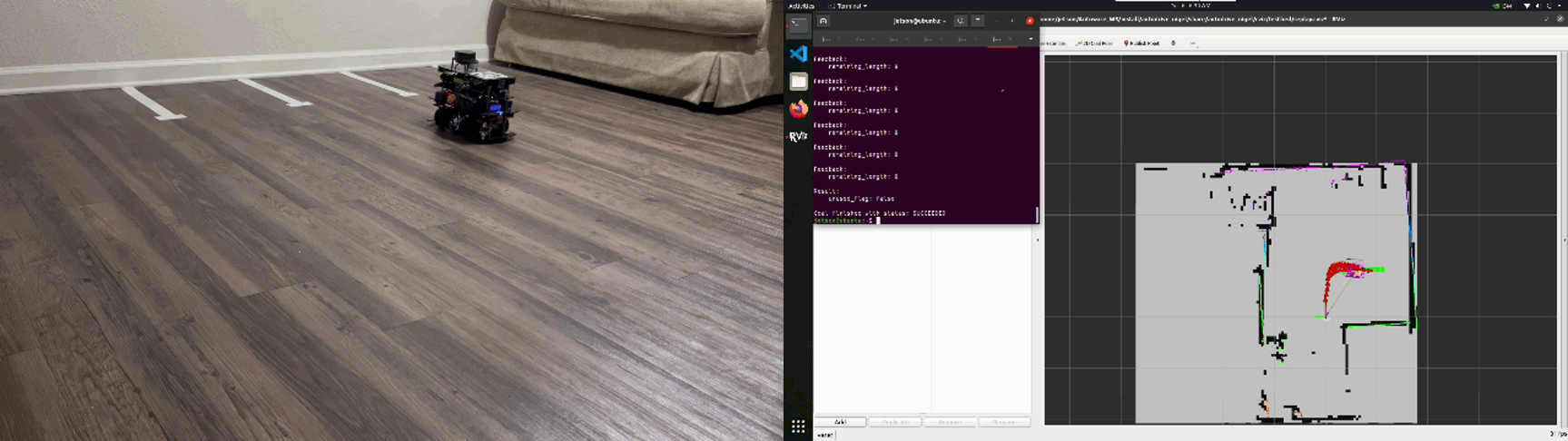}
    \caption{AutoDRIVE-Nigel-Autoware integration for autonomous valet parking ODD: Real-world demo for tracking the pre-recorded trajectory autonomously.}
    \label{fig: figure24}
\end{figure}

\hypertarget{Hunter SE}{%
\subsection{Hunter SE}\label{Hunter SE}}

\begin{figure}[H]
    \centering
    \includegraphics[width=\linewidth]{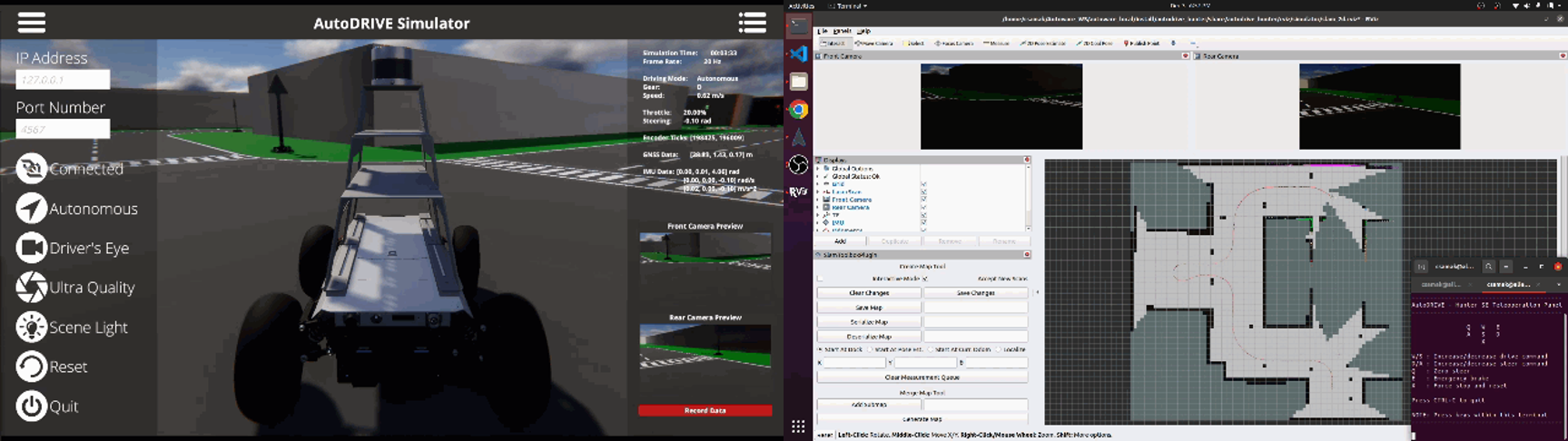}
    \caption{AutoDRIVE-HunterSE-Autoware integration for autonomous valet parking ODD: AutoDRIVE Simulator demo for mapping an unknown environment as a 2D binary occupancy grid (BOG) using manual teleoperation.}
    \label{fig: figure25}
\end{figure}

\begin{figure}[H]
    \centering
    \includegraphics[width=\linewidth]{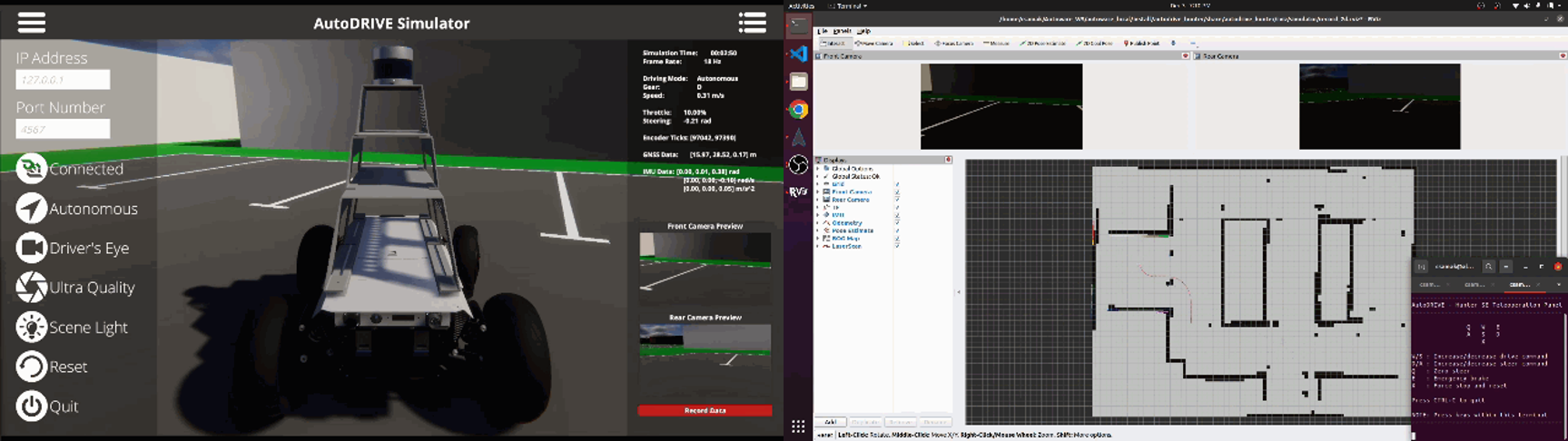}
    \caption{AutoDRIVE-HunterSE-Autoware integration for autonomous valet parking ODD: AutoDRIVE Simulator demo for recording a trajectory within the 2D binary occupancy grid (BOG) map using manual teleoperation.}
    \label{fig: figure26}
\end{figure}

\begin{figure}[H]
    \centering
    \includegraphics[width=\linewidth]{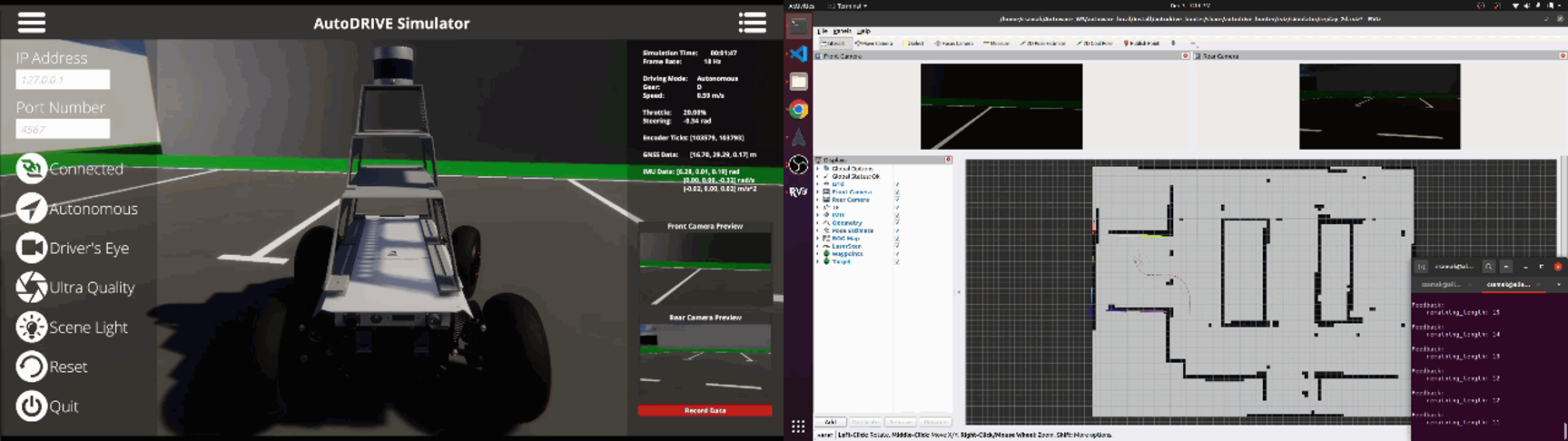}
    \caption{AutoDRIVE-HunterSE-Autoware integration for autonomous valet parking ODD: AutoDRIVE Simulator demo for tracking the pre-recorded trajectory autonomously within the 2D binary occupancy grid (BOG) map.}
    \label{fig: figure27}
\end{figure}

\begin{figure}[H]
    \centering
    \includegraphics[width=\linewidth]{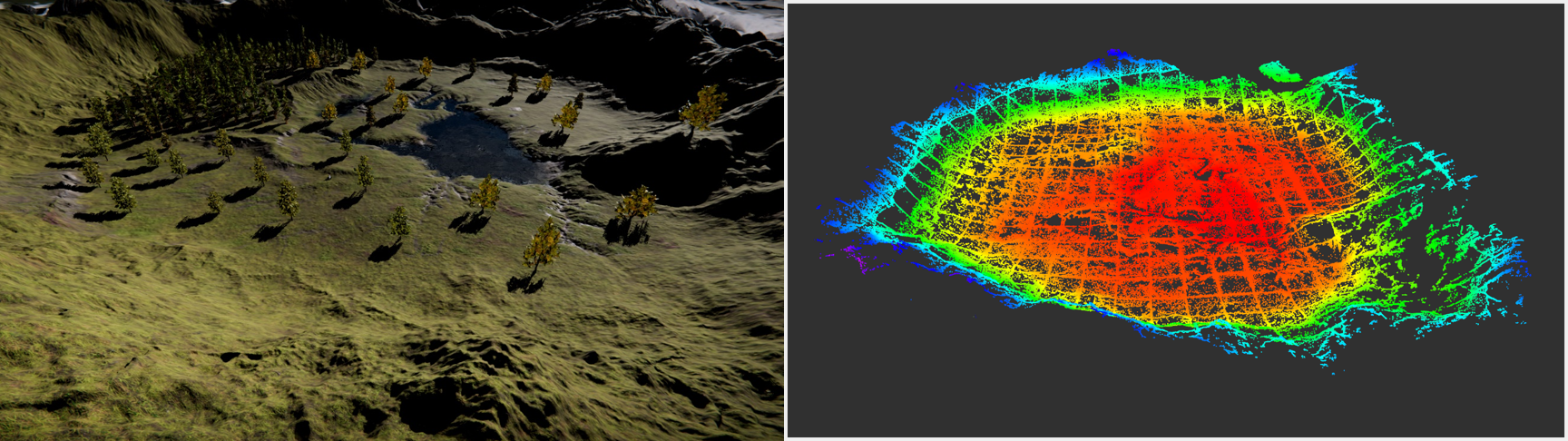}
    \caption{AutoDRIVE-HunterSE-Autoware integration for off-road exploration ODD: Generating and exporting 3D point cloud data (PCD) map directly from within AutoDRIVE Simulator using manual teleoperation.}
    \label{fig: figure28}
\end{figure}

\begin{figure}[H]
    \centering
    \includegraphics[width=\linewidth]{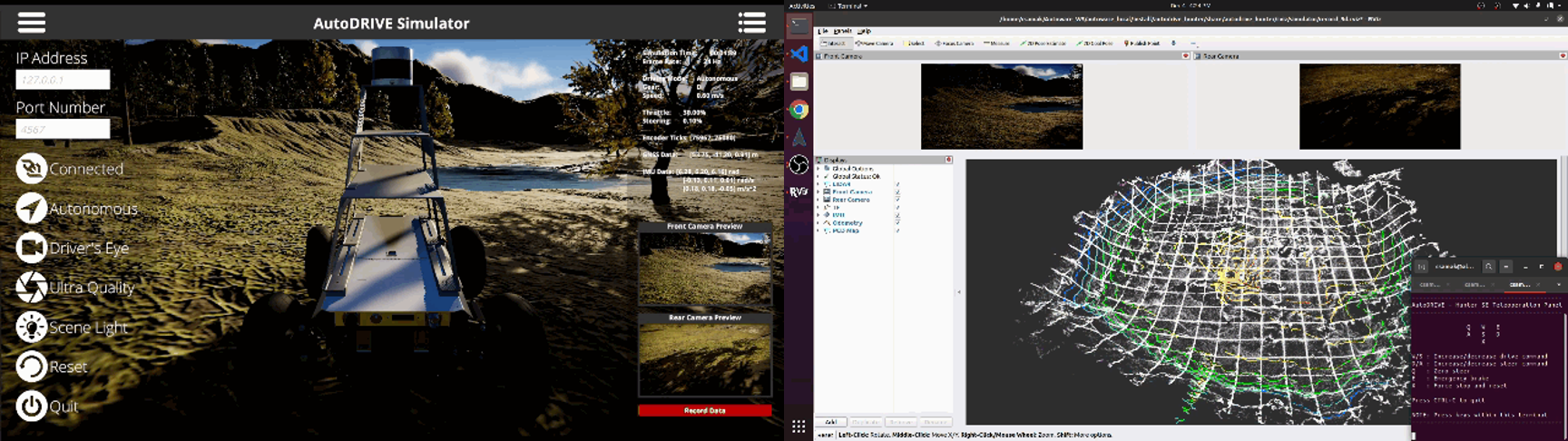}
    \caption{AutoDRIVE-HunterSE-Autoware integration for off-road exploration ODD: AutoDRIVE Simulator demo for recording a trajectory within the 3D point cloud data (PCD) map using manual teleoperation.}
    \label{fig: figure29}
\end{figure}

\begin{figure}[H]
    \centering
    \includegraphics[width=\linewidth]{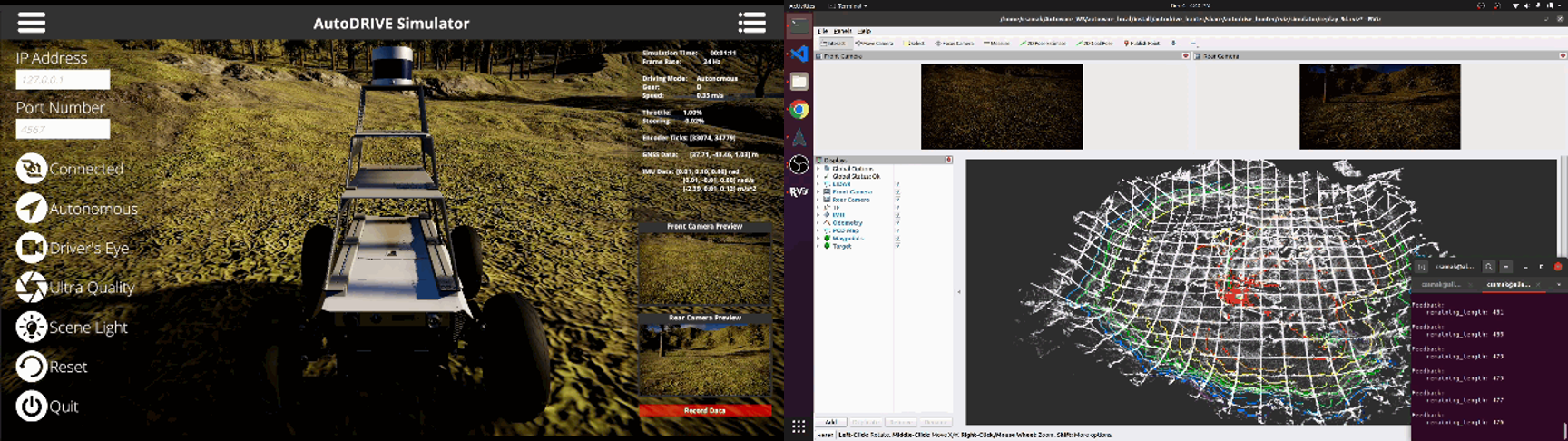}
    \caption{AutoDRIVE-HunterSE-Autoware integration for off-road exploration ODD: AutoDRIVE Simulator demo for tracking the pre-recorded trajectory autonomously within the 3D point cloud data (PCD) map.}
    \label{fig: figure30}
\end{figure}

\hypertarget{OpenCAV}{%
\subsection{OpenCAV}\label{OpenCAV}}

\begin{figure}[H]
    \centering
    \includegraphics[width=\linewidth]{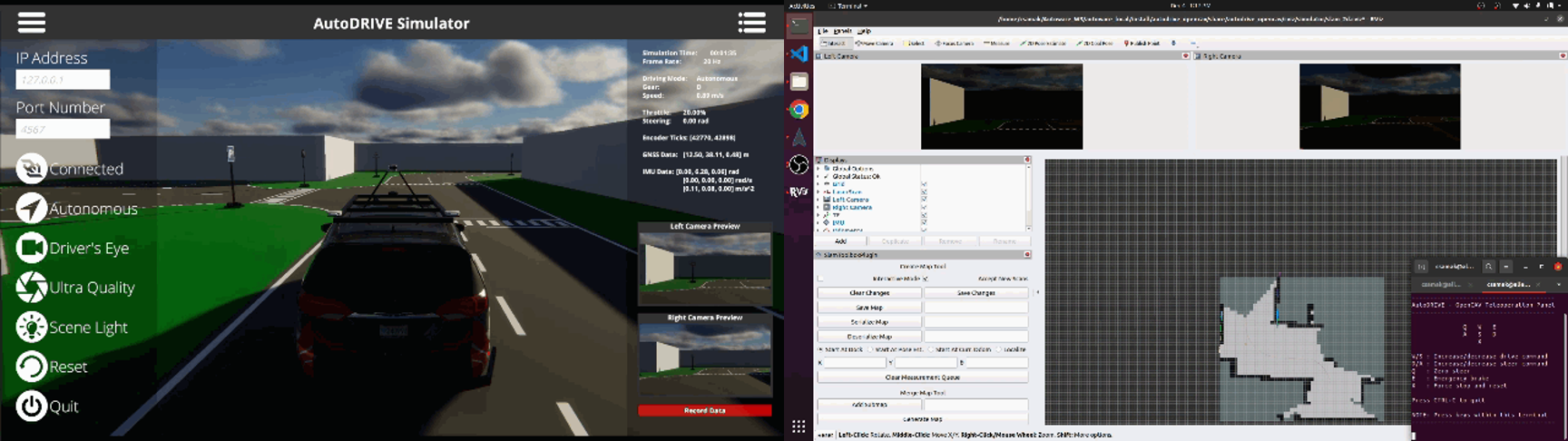}
    \caption{AutoDRIVE-OpenCAV-Autoware integration for autonomous valet parking ODD: AutoDRIVE Simulator demo for mapping an unknown environment as a 2D binary occupancy grid (BOG) using manual teleoperation.}
    \label{fig: figure31}
\end{figure}

\begin{figure}[H]
    \centering
    \includegraphics[width=\linewidth]{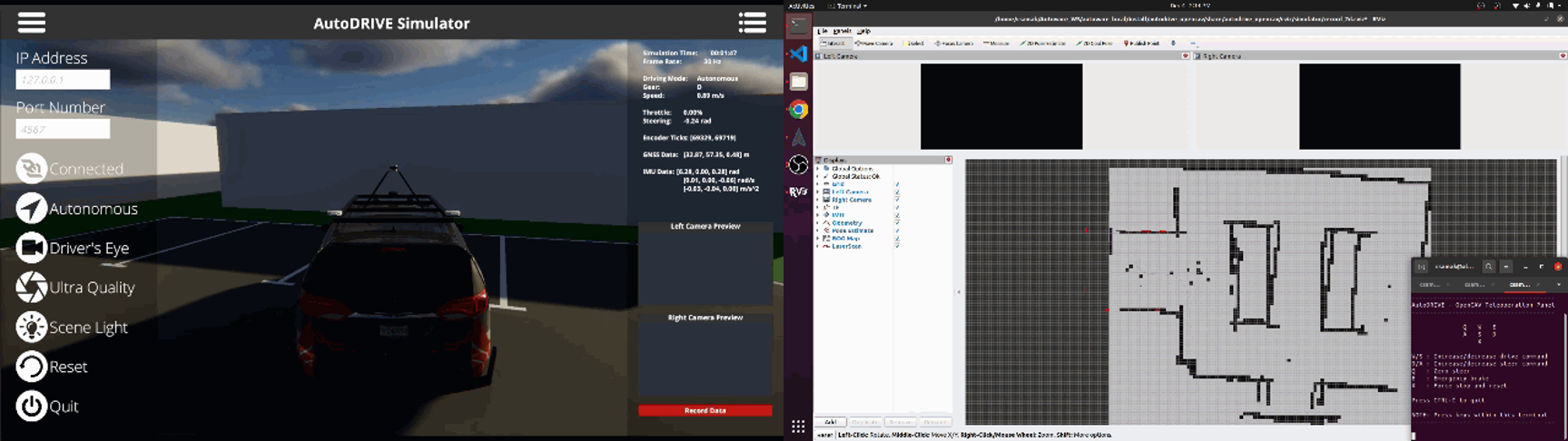}
    \caption{AutoDRIVE-OpenCAV-Autoware integration for autonomous valet parking ODD: AutoDRIVE Simulator demo for recording a trajectory within the 2D binary occupancy grid (BOG) map using manual teleoperation.}
    \label{fig: figure32}
\end{figure}

\begin{figure}[H]
    \centering
    \includegraphics[width=\linewidth]{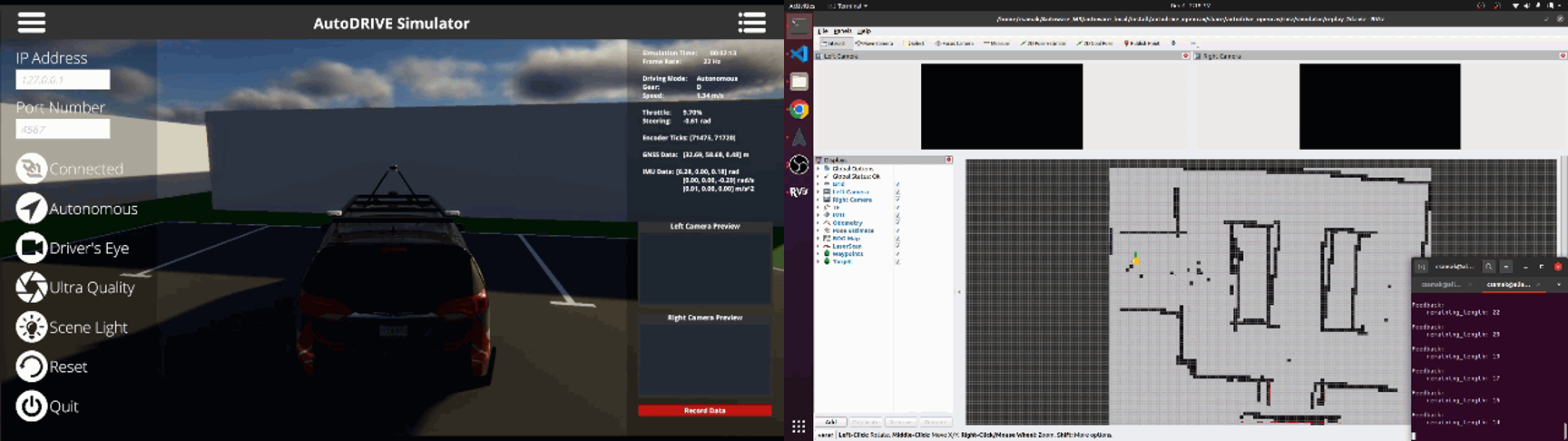}
    \caption{AutoDRIVE-OpenCAV-Autoware integration for autonomous valet parking ODD: AutoDRIVE Simulator demo for tracking the pre-recorded trajectory autonomously within the 2D binary occupancy grid (BOG) map.}
    \label{fig: figure33}
\end{figure}

\begin{figure}[H]
    \centering
    \includegraphics[width=\linewidth]{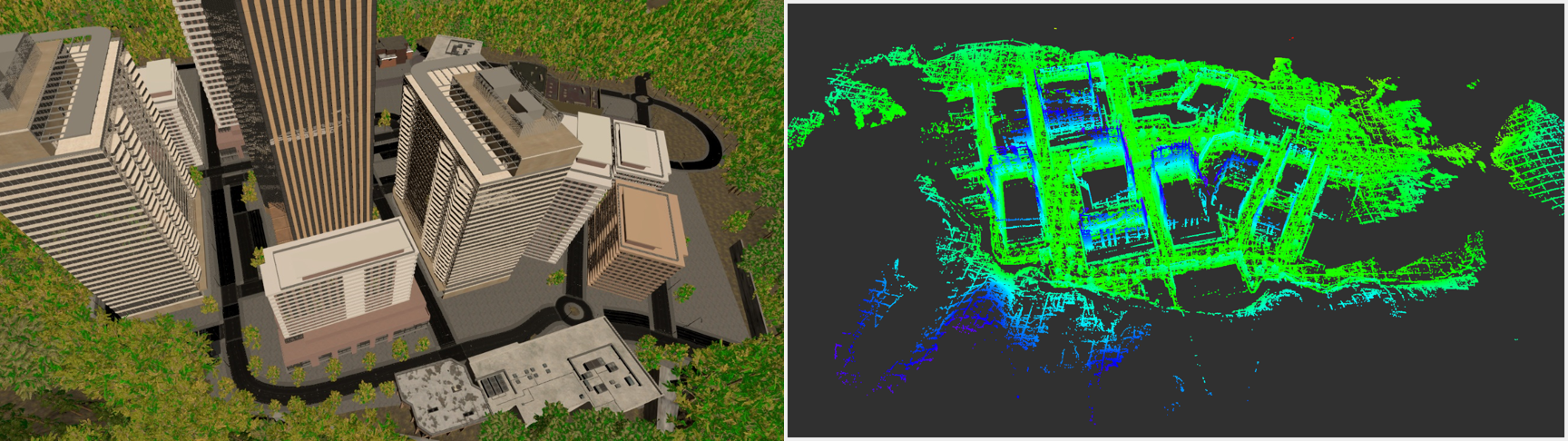}
    \caption{AutoDRIVE-OpenCAV-Autoware integration for autonomous valet parking ODD: Generating and exporting 3D point cloud data (PCD) map directly from within AutoDRIVE Simulator using manual teleoperation.}
    \label{fig: figure34}
\end{figure}

\begin{figure}[H]
    \centering
    \includegraphics[width=\linewidth]{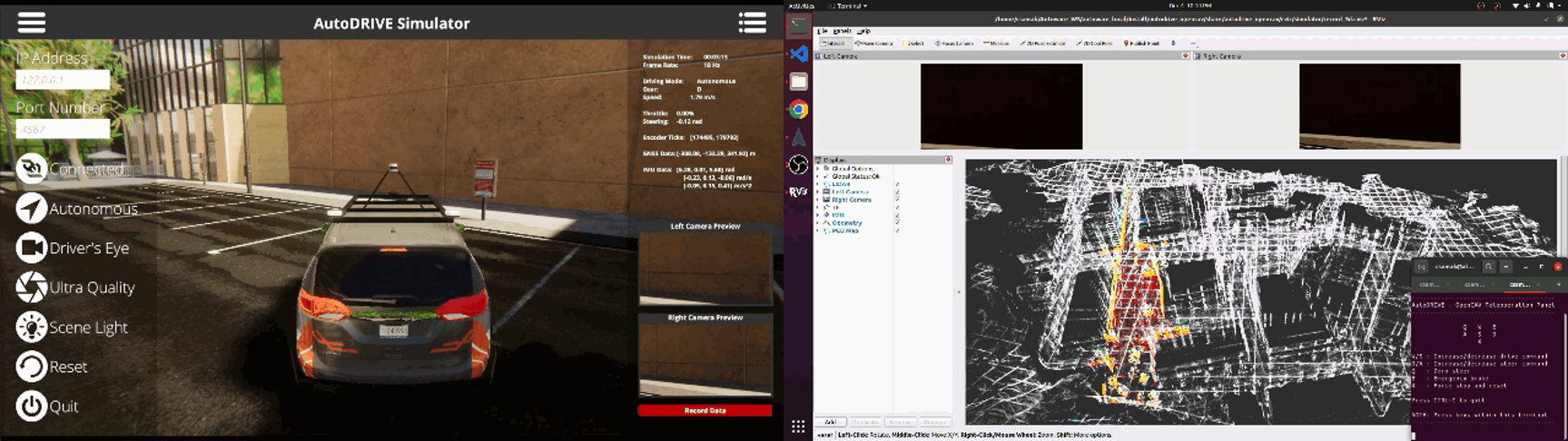}
    \caption{AutoDRIVE-OpenCAV-Autoware integration for autonomous valet parking ODD: AutoDRIVE Simulator demo for recording a trajectory within the 3D point cloud data (PCD) map using manual teleoperation.}
    \label{fig: figure35}
\end{figure}

\begin{figure}[H]
    \centering
    \includegraphics[width=\linewidth]{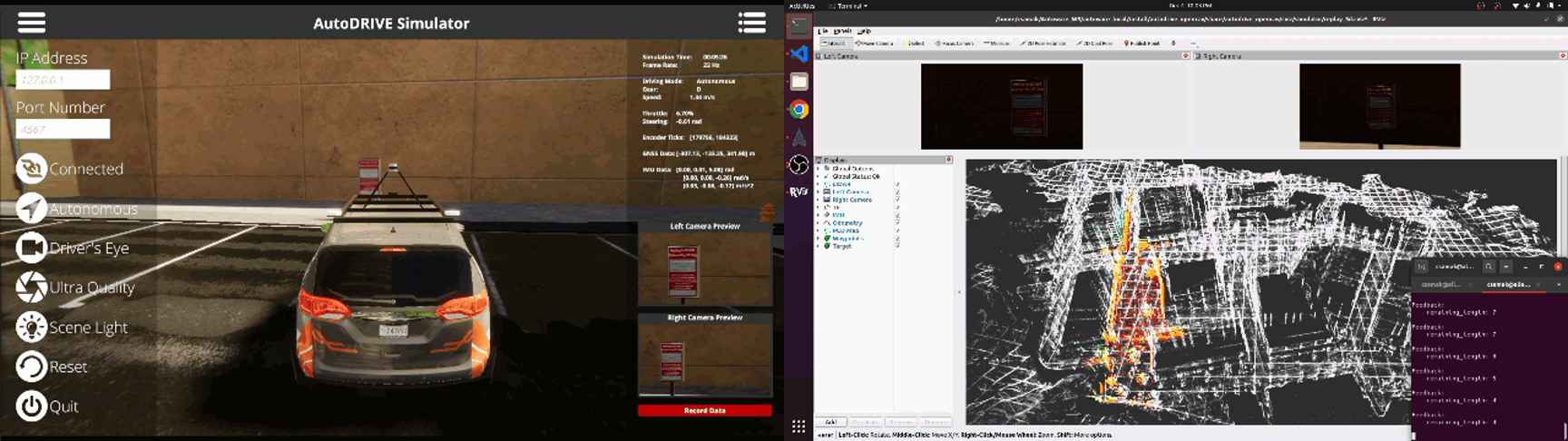}
    \caption{AutoDRIVE-OpenCAV-Autoware integration for autonomous valet parking ODD: AutoDRIVE Simulator demo for tracking the pre-recorded trajectory autonomously within the 3D point cloud data (PCD) map.}
    \label{fig: figure36}
\end{figure}

\pagebreak
\hypertarget{Concluding Remarks}{%
\section{Concluding Remarks}\label{Concluding Remarks}}

This work investigated the development of autonomy-oriented digital twins of vehicles across different scales and configurations to help support the streamlined development and deployment of Autoware Core/Universe stack, using a unified real2sim2real toolchain. Particularly, the core deliverable for this project was to integrate the Autoware stack with AutoDRIVE Ecosystem in order to demonstrate the end-to-end task of mapping an unknown environment, recording a trajectory within the mapped environment, and autonomously tracking the pre-recorded trajectory to achieve the desired objective. This work discussed the development of vehicle and environment digital twins using AutoDRIVE Ecosystem, along with various application programming interfaces (APIs) and human-machine interfaces (HMIs) to connect with the same, followed by a detailed section on AutoDRIVE-Autoware integration. It is worth mentioning that in addition to several Autoware deployment demonstrations, this study described the first-ever off-road deployment of the Autoware stack, thereby expanding its operational design domain (ODD) beyond on-road autonomous navigation.

\hypertarget{Novel Contributions}{%
\subsection{Novel Contributions}\label{Novel Contributions}}

\begin{figure}[t]
    \centering
    \includegraphics[width=\linewidth]{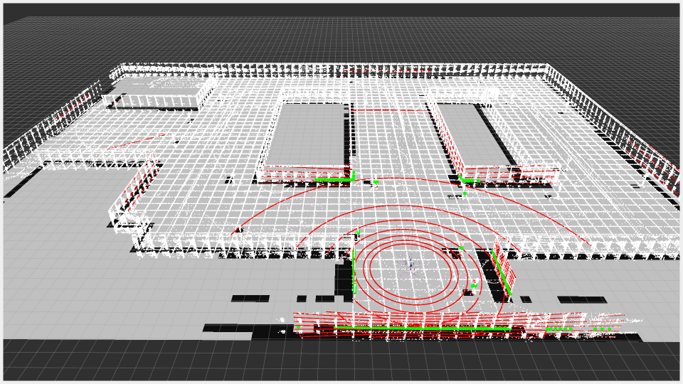}
    \caption{Working with a lightweight surrogate map of the environment by converting 3D LIDAR point cloud (\textcolor{red}{red}) to 2D laser scan (\textcolor{green}{green}) and applying planar mapping/localization stack for simplistic scenarios. Notice the PCD map (\textcolor{black!25}{white}) and BOG map (\textcolor{black!75}{gray}) overlaid on top of each other.}
    \label{fig: figure37}
\end{figure}

It is noteworthy to highlight that multiple high-level system configurations have been implemented within the original Autoware stack to ensure the effective operation of autonomous vehicles. Firstly, a trajectory looping criterion has been integrated to determine whether the very first waypoint in the reference trajectory should become the target waypoint for the controller upon successful tracking of the last waypoint in the reference trajectory. This feature is pivotal in shaping a ``safe termination'' behavior for the ego vehicle at the conclusion of the reference trajectory, particularly crucial in applications like autonomous valet parking. Additionally, the tolerance on the termination criteria is designed to enable the ego vehicle to smoothly come to a safe stop, avoiding unnecessary aggressive over-corrections or overshooting beyond the safe space. Secondly, an operational control mode has been introduced, allowing selective engagement of the ego vehicle in various modes of operation, including options of a simplistic gym environment, high-fidelity simulation environment, pure testbed (real-world hardware), or within a true digital twin framework leveraging the high-fidelity simulator in conjunction with the testbed.

Another distinctive contribution to the project lies in the creation of numerous custom meta-packages within the Autoware stack, addressing the variability in inputs, outputs, and configurations of perception, planning, and control modules across different vehicle platforms. This strategic approach ensures the project's overall cleanliness and organization. Each custom meta-package is designed to accommodate specific perception, planning, and control algorithms tailored to different vehicles within independent individual packages, thereby effectively managing diverse input and output information. Additionally, a dedicated meta-package has been established to streamline the handling of various vehicles within the AutoDRIVE Ecosystem utilized in this project, namely Nigel, F1TENTH, Hunter SE, and OpenCAV. Each individual package associated with a specific vehicle encompasses vehicle-specific parameter description configuration files for perception, planning, and control algorithms, along with map files, RViz configuration files, API program files, teleoperation program files, and user-friendly launch files, thereby ensuring a systematic and efficient management of project components.

Finally, in the case of mid and full-scale autonomous vehicles employing 3D LIDARs for exteroceptive perception of the environment, we have set up an end-to-end demonstration of converting the 3D point cloud data to a 2D laser scan datatype and operating on it for mapping and localization to aid in ``lightweight'' autonomous navigation (Fig. \ref{fig: figure37}). Particularly, this approach eradicates the need for utilizing complex algorithms for mapping and localization using 3D point cloud datatype. Instead, the converted 2D laser scan requiring much less computational overhead can be used with 2D mapping and localization algorithms widely adopted in the domain of autonomous mobile robots thereby reducing the computational burden from the algorithm complexity perspective as well. However, it is worth mentioning that this approach is only recommended when the environment is highly structured and loss of information from the third dimension creates little to no effect on the interpretation of environmental features based on the converted data.

\hypertarget{Challenges Faced}{%
\subsection{Challenges Faced}\label{Challenges Faced}}

\begin{figure}[t]
    \centering
    \includegraphics[width=\linewidth]{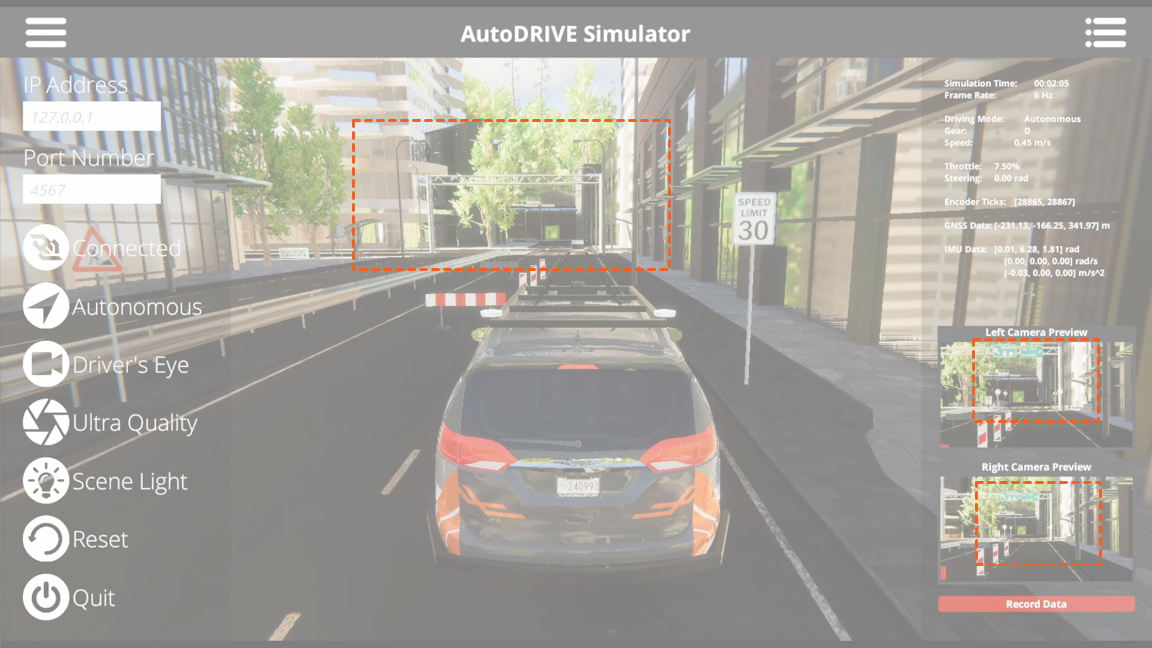}
    \caption{LOD culling gradually degrades the environmental details as they move further away from the scene camera. However, it does not affect any of the AV camera sensor(s). Notice the street signs at a distance culled from the scene camera, but visible in the AV camera sensor(s).}
    \label{fig: figure38}
\end{figure}

In the context of small-scale vehicles, the Autoware build process poses a significant challenge due to its prolonged duration (>8 hours) as well as its huge memory (RAM) requirement thereby impacting the efficiency of the development workflow. This problem is further aggravated since most small-scale autonomous vehicle platforms are battery-operated (with limited battery life) and therefore it is necessary to disconnect all the electrically incompatible peripherals from the onboard computer and use shore power during the installation and setup phase. Additionally, conflicts arising from different ROS distributions and parallelly sourced workspaces further complicate the integration and deployment of the Autoware stack, necessitating careful resolutions such as proper isolation of environment variables to ensure seamless operation.

For mid and full-scale vehicles, the project encounters challenges related to the management of 3D LIDAR point clouds, requiring a robust approach to handle the complexity and volume of data generated. Balancing the level of detail (LOD) in simulation versus perception is a critical consideration, introducing challenges in optimizing simulation fidelity without compromising real-world accuracy (Fig. \ref{fig: figure38}). Moreover, maneuvering in tight spaces poses a distinct challenge, demanding precise control algorithms and navigation strategies to ensure the safe and effective operation of full-scale vehicles within constrained environments. Addressing these challenges is crucial for the successful development and deployment of autonomous systems for both small and full-scale vehicles.

\hypertarget{Future Plans}{%
\subsection{Future Plans}\label{Future Plans}}

\begin{figure}[h]
     \centering
     \begin{subfigure}[b]{0.49\linewidth}
         \centering
         \includegraphics[width=\linewidth]{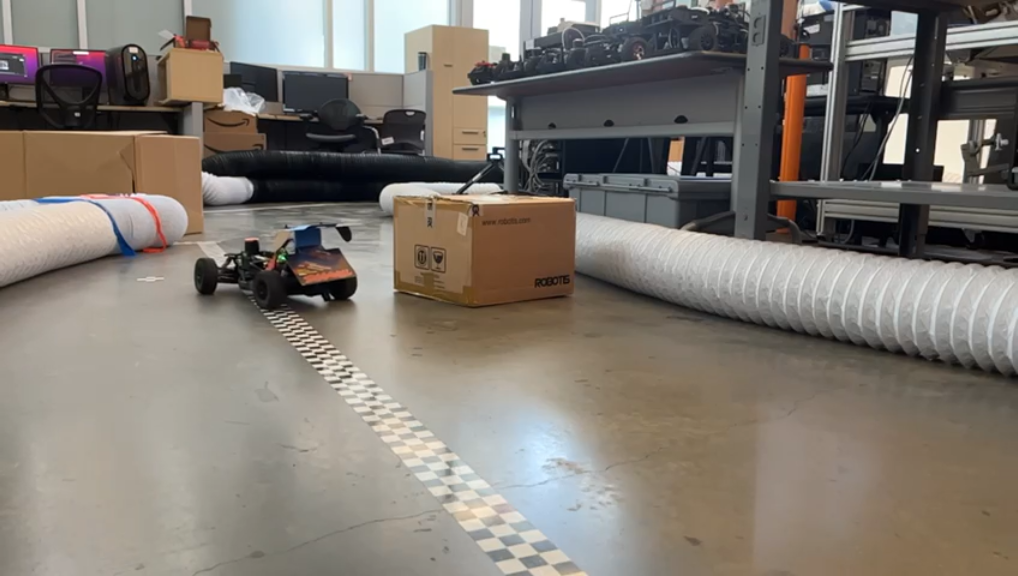}
         \caption{F1TENTH}
         \label{fig39a}
     \end{subfigure}
     \hfill
     \begin{subfigure}[b]{0.49\linewidth}
         \centering
         \includegraphics[width=\linewidth]{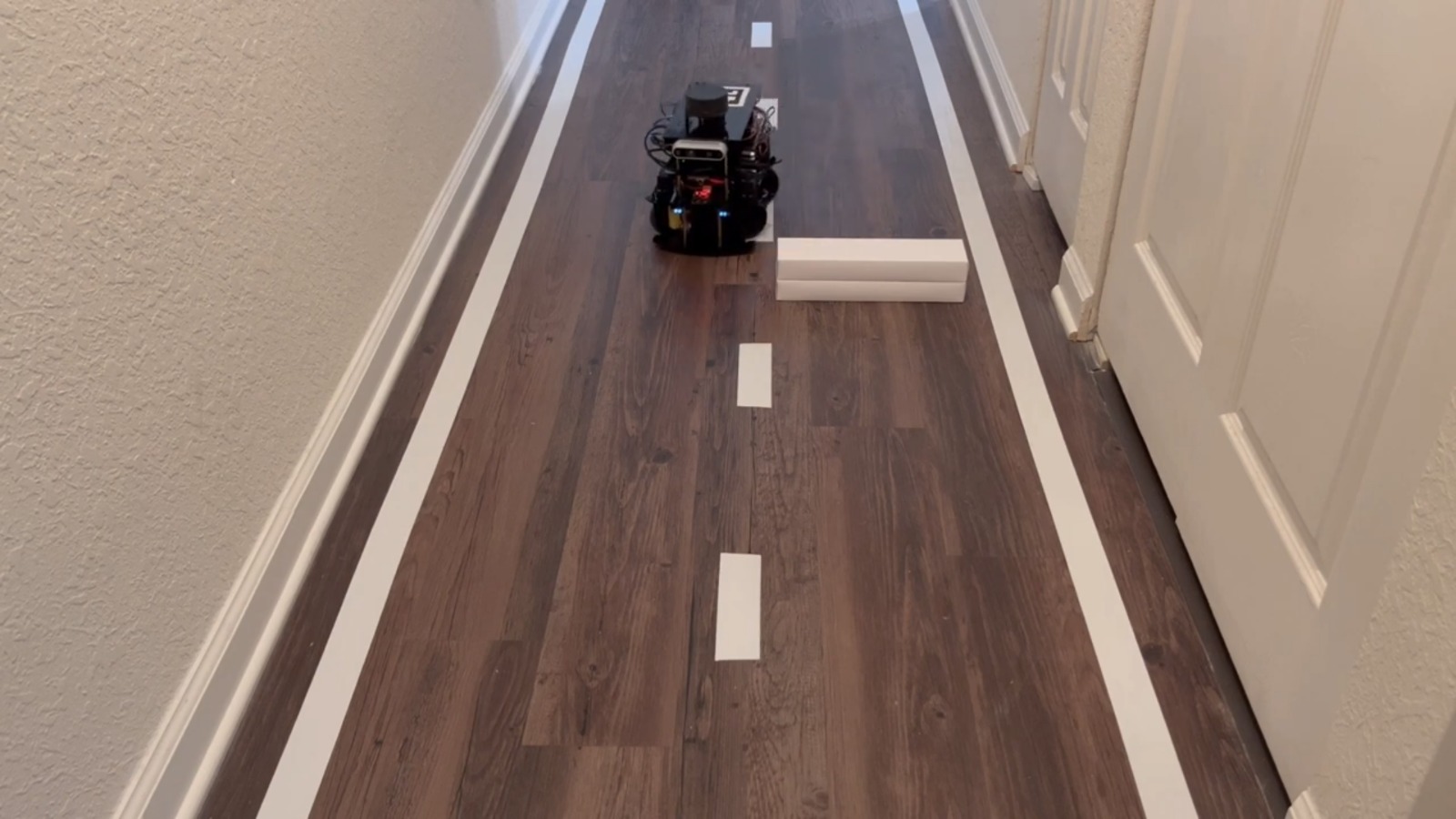}
         \caption{Nigel}
         \label{fig39b}
     \end{subfigure}
     \caption{F1TENTH and Nigel demonstrating their dynamic collision avoidance capability independent of Autoware stack. Integrating this functionality within the Autoware deployment is a future plan.}
    \label{figure39}
\end{figure}

In the future trajectory of the project, several key developments are envisioned across different scales of vehicles. For small-scale vehicles, the focus lies on advancing the capabilities through multi-agent deployment, enhancing the system's adaptability to collaborative and/or competitive scenarios. Additionally, the implementation of dynamic re-planning strategies (Fig. \ref{figure39}) is imperative to address real-time environmental changes and optimize the overall navigation efficiency (especially exploiting the redundancy in 4WD4WS Nigel by solving multi-objective optimization a.k.a. Pareto optimization problems).

Mid-scale vehicles will see a crucial step forward with the sim2real configuration and setup of the Autoware stack, ensuring a seamless transition of developed algorithms from simulation to real-world deployment. This transfer is pivotal for validating the system's performance in actual operational environments. Furthermore, dynamic re-planning remains a priority for mid-scale vehicles as well, allowing the system to respond dynamically to changing scenarios and optimize path planning strategies.

For full-scale vehicles, the sim2real configuration of Autoware stack continues to be a significant focus, emphasizing the importance of validating the system's autonomy in real-world scenarios. Dynamic re-planning remains another critical aspect for adapting to dynamic environments and ensuring the safety and efficiency of autonomous operations.

These future directions planned for expanding this project even further collectively aim to enhance the functionality, adaptability, and safety of autonomous vehicles across different scales and operational design domains whilst running the Autoware stack as an open-source medium or framework for autonomous vehicle software development.

\hypertarget{Supplemental Materials}{%
\subsection{Supplemental Materials}\label{Supplemental Materials}}

All source files including codebase, project proposal, presentation slides and technical writeups used for the completion of various aspects of the capstone project are openly available on GitHub: \href{https://github.com/Tinker-Twins/Scaled-Autonomous-Vehicles}{https://github.com/Tinker-Twins/Scaled-Autonomous-Vehicles}. We have intentionally avoided mixing the codebase for Autoware deployments within this repository for cleaner organization.

The codebase for AutoDRIVE-Autoware integration has been publicly released and is openly available on GitHub. Following is a link to our repository: \href{https://github.com/Tinker-Twins/AutoDRIVE-Autoware}{https://github.com/Tinker-Twins/AutoDRIVE-Autoware}. It is worth mentioning that this repository does not host any source files pertaining to AutoDRIVE Ecosystem itself. AutoDRIVE is openly accessible through another GitHub repository of ours: \href{https://github.com/Tinker-Twins/AutoDRIVE}{https://github.com/Tinker-Twins/AutoDRIVE}. Additionally, AutoDRIVE Ecosystem has its own GitHub Organization, which hosts additional repositories: \href{https://github.com/AutoDRIVE-Ecosystem}{https://github.com/AutoDRIVE-Ecosystem}

Although the primary aim of using Git repositories for this project was to enable version control and collaborative development, this also gave us the opportunity to document all the progress using markdown files (\texttt{README.md}). Despite challenges and scarcity of time, we invested a significant amount of time and effort in building and maintaining our repositories throughout this project. We hope that this repository will help the readers better understand our efforts, outcomes and learnings from this project and potentially serve as documentation for us (or others) when trying to explore similar approaches.

\pagebreak
}

\printbibliography

\end{document}